\documentclass[a4paper,12pt]{article}
\usepackage{fullpage}
\usepackage[noend]{algorithmic}
\usepackage{amsmath}
\usepackage{color}
\usepackage[nocenter]{cwpuzzle}
\usepackage{mathtools}
\usepackage{tikz} \usetikzlibrary{arrows,automata}
\usepackage{hyperref}

\newcommand{\EuroControl}{EUROCONTROL}

\newcommand{\red}[1]{{\color{red}#1}}

\newcommand{\Constraint}[1]{\textsc{#1}}
\newcommand{\AllDifferent}{\Constraint{AllDifferent}}
\newcommand{\Automaton}{\Constraint{Automaton}}
\newcommand{\Balanced}{\Constraint{Balanced}}
\newcommand{\Bounded}{\Constraint{Bounded}}
\newcommand{\Compact}{\Constraint{Compact}}
\newcommand{\Connected}{\Constraint{Connected}}

\newcommand{\Deviation}{\Constraint{Deviation}}
\newcommand{\Linear}{\Constraint{Linear}}
\newcommand{\NonBorder}{\Constraint{NonBorder}}
\newcommand{\Spread}{\Constraint{Spread}}
\newcommand{\StretchSum}{\Constraint{StretchSum}}

\newcommand{\Arc}[2]{#1~\Set{#2}}  
\newcommand{\CurrentAssignment}{\alpha}
\newcommand{\CurrentValue}[1]{\CurrentAssignment(#1)}
\newcommand{\Domain}[1]{\text{dom}(#1)}
\newcommand{\Eq}{$\Set{k \IsAssigned \Val{\First{V}}}$}
\newcommand{\HigherOrder}[1]{\red{\left[{\color{black}#1}\right]}} 
\newcommand{\NewAssignment}{\alpha'}
\newcommand{\NewValue}[1]{\NewAssignment(#1)}
\newcommand{\Probe}[1]{\Delta(#1)}
\newcommand{\ProbeVar}[2]{\delta(#1,#2)}  
\newcommand{\Reifies}[2]{#1\Leftrightarrow#2}
\newcommand{\Violation}{\text{violation}}
\newcommand{\VarViolation}[1]{\Violation(#1)}

\newcommand{\In}{\textbf{in~}}
\newcommand{\IsAssigned}{\coloneqq}
\newcommand{\Let}{\textbf{let~}}
\newcommand{\LetIn}[2]{\Let #1 \\ \In #2}

\newcommand{\Bord}{\mathit{Border}}
\newcommand{\Border}[1]{\Bord(#1)}
\newcommand{\CCs}{\mathit{CCs}}
\newcommand{\Colour}{\mathit{Colour}}
\newcommand{\Col}[1]{\Colour(#1)}
\newcommand{\Colours}{\mathit{Colours}}
\newcommand{\NCC}{\mathit{NCC}}       
\newcommand{\NCCcolour}[1]{\NCC(#1)}  
\newcommand{\Sig}[1]{\mathit{Sig}(#1)}
\newcommand{\Stretch}{\mathit{Stretch}}
\newcommand{\StretchVertex}[1]{\Stretch(#1)}
\newcommand{\Value}{\mathit{Value}}
\newcommand{\Val}[1]{\Value(#1)}
\newcommand{\Vol}[1]{\Volume(#1)}

\newcommand{\AbsValue}[1]{\left\lvert#1\right\rvert}
\newcommand{\Adjacent}[1]{\text{Adj}(#1)}
\newcommand{\Area}{\text{Area}}  
\newcommand{\Cardinality}[1]{\left\lvert#1\right\rvert}
\newcommand{\Ceiling}[1]{\left\lceil#1\right\rceil}
\newcommand{\ColourGraph}{\mathit{ColourGraph}}
\newcommand{\EmptySet}{\varnothing}
\newcommand{\enveloped}[1]{#1_{\bot}}
\newcommand{\facet}[1]{\facetfunname(#1)}
\newcommand{\facetfunname}{\partial}
\newcommand{\facetset}{F}
\newcommand{\First}[1]{\text{first}(#1)}
\newcommand{\Floor}[1]{\left\lfloor#1\right\rfloor}
\newcommand{\Iff}{\Leftrightarrow}
\newcommand{\Implies}{\Rightarrow}
\newcommand{\Intersect}{\cap}
\newcommand{\Nat}{\mathbb{N}}
\newcommand{\Oh}[1]{\mathcal{O}\left(#1\right)}
\newcommand{\Pred}[1]{\text{pred}(#1)}
\newcommand{\RelOp}{\mathit{RelOp}}
\newcommand{\Sequence}[1]{\left[#1\right]}
\newcommand{\Set}[1]{\left\{#1\right\}}
\newcommand{\Setcomp}[2]{\Set{#1 \SuchThat #2}}
\newcommand{\Succ}[1]{\text{succ}(#1)}
\newcommand{\SuchThat}{\mid}
\newcommand{\Tuple}[1]{\left\langle#1\right\rangle}
\newcommand{\Union}{\cup}
\newcommand{\Volume}{\text{Volume}}

\newcommand{\Plan}{\mathit{Plan}}
\newcommand{\regions}{\mathit{Regions}}
\newcommand{\workloaddomain}{\mathit{Workloads}}
\newcommand{\workloadfun}{\text{Workload}}
\newcommand{\workloadcombine}{\text{C}}

\begin{document}

\title{\textbf{Propagators and Violation Functions for \\
    Geometric and Workload Constraints \\
    Arising in Airspace Sectorisation \thanks{This work has been
      co-financed by the European Organisation for the Safety of Air
      Navigation (\EuroControl) under its Research Grants programme
      (contract 08-1214447-C).  The content of the work does not
      necessarily reflect the official position of \EuroControl\ on
      the matter.}}}

\author{Pierre Flener and Justin Pearson \\
  Department of Information Technology \\
  Uppsala University, Box 337, SE -- 751 05 Uppsala, Sweden \\
  \url{Firstname.Surname@it.uu.se}}

\date{Revision of 29 January 2014}

\maketitle

\begin{abstract}\noindent
  Airspace sectorisation provides a partition of a given airspace into
  sectors, subject to geometric constraints and workload constraints,
  so that some cost metric is minimised.  We make a study of the
  constraints that arise in airspace sectorisation.  For each
  constraint, we give an analysis of what algorithms and properties
  are required under systematic search and stochastic local search.
\end{abstract}

\section{Introduction}

We continue our work in \cite{Jaegare:MSc11,ASTRA:ATM13} on applying
constraint programming (see Section~\ref{sect:cp}) to airspace
sectorisation (see Section~\ref{sect:secto}).  In that work, we used
existing constraints or implemented new ones in an ad hoc fashion just
to solve the problem at hand, without much theoretical analysis of the
constraints and their underlying algorithms.

We give a complete theoretical analysis of constraints that arise in
airspace sectorisation.  For each constraint, we analyse what
propagation is possible under systematic search.  Under stochastic
local search, we give efficient algorithms for maintaining constraint
and variable violations, as well as efficient algorithms for probing
the effect of local search moves; such algorithms are necessary if
stochastic local search is to be done efficiently.

The remainder of this report is structured as follows.  Towards making
it self-contained, we first give in Section~\ref{sect:secto} a brief
overview on airspace sectorisation, and in Section~\ref{sect:cp} a
brief tutorial on constraint programming, especially about its core
concept: a constraint is a reusable software component that can be
used declaratively, when modelling a combinatorial problem, and
procedurally, when solving the problem.  Experts on airspace
sectorisation or CP may safely skip these sections.  Next, in
Section~\ref{sect:math}, we lay the mathematical foundation for
specifying, in Section~\ref{sect:geo}, the constraints identified
above to occur commonly in airspace sectorisation problems Finally, in
Section~\ref{sect:concl}, we summarise our contributions and identify
avenues for future work.

\section{Background: Airspace Sectorisation}
\label{sect:secto}

Airspace \emph{sectorisation} provides a partition of a given airspace
into a given (or upper-bounded, or minimal) number of control sectors,
subject to geometric constraints and workload constraints, so that
some cost metric is minimised.  The entire material of this section is
a suitably condensed version of Section~2 of our
report~\cite{ASTRA:sectorisation:survey}, where we have surveyed the
algorithmic aspects of methods for automatic airspace sectorisation,
for an intended readership of experts on air traffic management.

In this report, we focus on \emph{region-based} models of airspace
sectorisation, where the airspace is initially partitioned into some
kind of regions that are smaller than the targeted sectors, so that
the combinatorial problem of partitioning these regions in principle
needs no geometric post-processing step.  Sectorisation then starts
from regions of (any combination of) the following granularities: a
mesh of blocks of the same size and shape; ATC functional blocks
(AFBs); elementary sectors, namely the ones of the existing
sectorisation; control sectors; areas of specialisation (AOS); and air
traffic control centres (ATCC).

We assume in this report that the airspace sectorisation is computed
in \emph{three} dimensions, hence the ideas scale down to two
dimensions.  For simplifying the discussion, we assume without loss of
generality in this report that the number of sectors is a \emph{given}
constant, rather than upper-bounded or to be minimised.

Airspace sectorisation aims at satisfying some constraints.  The
following constraints have been found in the literature, so that a
subset thereof is chosen for a given tool:
\begin{itemize}
\item \emph{Balanced workload}: The workload of each sector must be
  within some given imbalance factor of the average across all
  sectors.
\item \emph{Bounded workload}: The workload of each sector must not
  exceed some upper bound.
\item \emph{Balanced size}: The size of each sector must be within
  some given imbalance factor of the average across all sectors.
\item \emph{Minimum dwell time}: Every flight entering a sector must
  stay within it for a given minimum amount of time (say two minutes),
  so that the coordination work pays off and that conflict management
  is possible.
\item \emph{Minimum distance}: Each existing trajectory must be inside
  each sector by a minimum distance (say ten nautical miles), so that
  conflict management is entirely local to sectors.
\item \emph{Convexity} of the  sectors.  Convexity can be in
  the usual \emph{geometric} sense, or \emph{trajectory-based} (no
  flight enters the same sector more than once), or more complex.
\item \emph{Connectedness}: A sector must be a contiguous portion of
  airspace and can thus not be fragmented into a union of unconnected
  portions of airspace.
\item \emph{Compactness}: A sector must have a geometric shape that is
  easy to keep in mind.
\item \emph{Non-jagged boundaries}: A sector must have a boundary that
  is not too jagged.
\end{itemize}
For each sector, there are three kinds of workload: the
\emph{monitoring workload}, the \emph{conflict workload}, and the
\emph{coordination workload}; the first two workloads occur inside the
sector, and the third one between the sector and an adjacent sector.
The quantitative definition of workload varies strongly between
papers.

Airspace sectorisation often aims at minimising some cost.  The
following costs have been found in the literature, so that a subset
thereof is combined into the cost function for a given tool, the
subset being empty if sectorisation is not seen as an optimisation
problem:
\begin{itemize}
\item \emph{Coordination cost}: The cost of the total coordination
  workload between the sectors must be minimised.
\item \emph{Transition cost}: The cost of switching from the old
  sectorisation to the new one must be minimised.
\item \emph{Workload imbalance}: The imbalance between the workload of
  the sectors must be minimised.
\item \emph{Number of sectors}: The number of sectors must be
  minimised.
\item \emph{Entry points}: The total number of entry points into the
  sectors must be minimised.
\item If any of the constraints above is soft, then there is the
  additional cost of minimising the number of violations of soft
  constraints.
\end{itemize}
Constraints are to be \emph{satisfied}, hence the existence or not of
a cost function whose value is to be \emph{optimised} does not affect
the design of a constraint, in the sense discussed in the following
section.  Hence we will no further discuss cost functions in this
report.

\section{Background: Constraint Programming}
\label{sect:cp}

First, we show how to model combinatorial problems at a very high
level of abstraction with the help of the declarative notion of
\emph{constraint} (Section~\ref{sect:model}).  One distinguishes
between \emph{satisfaction problems}, where there is no objective
function, and \emph{optimisation problems}, where there is an
objective function.  Then, we describe two methods for solving
combinatorial problems by a mixture of inference and search, upon
reusing algorithms that implement these constraints
(Section~\ref{sect:solve}).  The entire material of this section is a
condensed version of our tutorial in Section~2 of
\cite{ASTRA:KER:survey}.

\subsection{Problem Modelling via Constraints}
\label{sect:model}

\emph{Constraint programming} (CP) is a successful approach to the
modelling and solving of combinatorial problems.  Its core idea is to
capture islands of structure, also called combinatorial
sub-structures, that are commonly occurring within such problems and
to encapsulate declaratively their procedural inference algorithms by
designing specialised reusable software components, called
\emph{constraints}.

\begin{figure}[t]
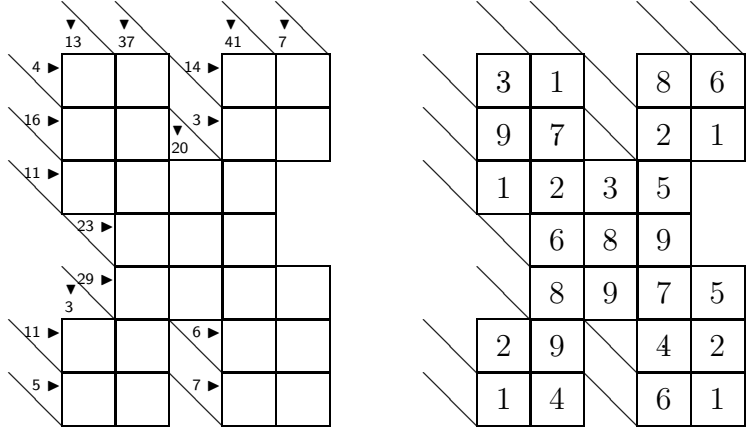

  \begin{center}
    \begin{Kakuro}{6}{9}
      |  -  |<:13> |<:37>|  -   |<:41>|<:7> | - |.
      |<4:> |  3   |  1  |<14:> |  8  |  6  | - |.
      |<16:>|  9   |  7  |<3:20>|  2  |  1  | - |.
      |<11:>|  1   |  2  |  3   |  5  |  -  | - |.
      |  -  |<23:> |  6  |  8   |  9  |  -  | - |.
      |  -  |<29:3>|  8  |  9   |  7  |  5  | - |.
      |<11:>|  2   |  9  |<6:>  |  4  |  2  | - |.
      |<5:> |  1   |  4  |<7:>  |  6  |  1  | - |.
      |  -  |  -   |  -  |  -   |  -  |  -  | - |.
    \end{Kakuro}
    ~~~~~~
    \PuzzleSolution
    \begin{Kakuro}{6}{9}
      |  -  |<:13> |<:37>|  -   |<:41>|<:7> | - |.
      |<4:> |* 3   |  1  |<14:> |  8  |  6  | - |.
      |<16:>|  9   |* 7  |<3:20>|  2  |  1  | - |.
      |<11:>|  1   |  2  |  3   |  5  |  -  | - |.
      |  -  |<23:> |  6  |* 8   |  9  |  -  | - |.
      |  -  |<29:3>|  8  |  9   |  7  |  5  | - |.
      |<11:>|  2   |  9  |<6:>  |* 4  |  2  | - |.
      |<5:> |  1   |  4  |<7:>  |  6  |  1  | - |.
      |  -  |  -   |  -  |  -   |  -  |  -  | - |.
    \end{Kakuro}
    \caption{A Kakuro puzzle (on the left) and its solution (on the
      right), both taken from the crossword package for \LaTeX\ by
      Gerd Neugebauer at \url{CTAN.org}.}
    \label{fig:kakuro}
  \end{center}
\end{figure}

As a running example, let us consider the Kakuro puzzle, with the
caveat that this is \emph{not} meant to imply that the discussed
techniques are limited to (small) puzzles.  See the left side of
Figure~\ref{fig:kakuro}: Each word of a Kakuro puzzle is made of
non-zero digits, under the two constraints that the letters of each
word are pairwise distinct and add up to the number to the left (for
horizontal words) or on top (for vertical words) of the word.  Hence
there is one decision variable for each letter of the puzzle, with the
integer set $\Set{1,2,\dots,9}$ as domain.  This is a satisfaction
problem, as there is no objective function.  The solution to this
puzzle is given on the right side of Figure~\ref{fig:kakuro}; note
that a Kakuro puzzle is always designed so as to have a unique
solution.

Beside the usual binary comparison constraints ($<$, $\leq$, $=$,
$\neq$, $\geq$, $>$), with arguments involving the usual binary
arithmetic operators ($+$, $-$, $\cdot$, etc), a large number of
\emph{combinatorial constraints} (usually called \emph{global
  constraints} in the literature), involving more complex structures
and a non-fixed number of arguments, have been identified.

For example, the $\AllDifferent(\Set{x_1,\dots,x_n})$ constraint
requires its $n$ decision variables~$x_i$ to take pairwise different
values.  This constraint is useful in many challenging real-life
problems, as well as for modelling the first constraint of the Kakuro
puzzle, as there is an $\AllDifferent$ constraint on each word.  Each
such constraint wraps a conjunction of $\frac{n \cdot (n-1)}{2}$
binary $x_i \neq x_j$ constraints into a single $n$-ary constraint;
this enables a more global view of the structure of the problem, which
is a prerequisite for solving it efficiently (as we shall see in
Section~\ref{sect:solve}).

As another example, the $\Linear(\Set{x_1,\dots,x_n},\RelOp,c)$
constraint requires the sum $x_1+\dots+x_n$ of the $n$ decision
variables~$x_i$ to be in relation $\RelOp$ with the constant $c$.
This constraint can be used for modelling the second constraint of the
Kakuro puzzle, as there is a $\Linear(\Set{x_1,\dots,x_n},=,s)$
constraint for each word $\Sequence{x_1,\dots,x_n}$ with required sum
$s$.  The generalised constraint
$\Linear(\Sequence{a_1,\dots,a_n},\Sequence{x_1,\dots,x_n},\RelOp,c)$
requires the \emph{weighted} sum $a_1 \cdot x_1 +\dots+ a_n \cdot x_n$
of the $n$ decision variables~$x_i$ to be in relation $\RelOp$ with
the constant $c$, where the weights $a_i$ are constants.

Hundreds of useful constraints have been identified and are described
in the \emph{Global Constraint Catalogue} \cite{GC-catalogue}, the
most famous ones being \Constraint{element} \cite{element},
\Constraint{cumulative} \cite{cumulative}, \Constraint{alldifferent}
\cite{alldifferent}, and \Constraint{cardinality} \cite{gcc}.

Constraints can be combined to \emph{model} a combinatorial problem in
a declarative and very high-level fashion.  For example, the Kakuro
puzzle can essentially be modelled as follows, assuming that a hint
for word $w$ with required sum $s$ is encoded as the pair
$\Tuple{w,s}$:
\begin{equation} \label{model:kakuro:naive}
  \begin{array}{c}
    \mathbf{for~each} \mathit{~hint~}
    \Tuple{\Sequence{x_1,\dots,x_n},s}: \\
    \AllDifferent(\Set{x_1,\dots,x_n}) \land
    \Linear(\Set{x_1,\dots,x_n},=,s)
  \end{array}
\end{equation}

For combinatorial problems, CP is \emph{not} limited to decision
variables whose domains are finite sets of integers.  For instance,
domains can be finite sets of \emph{finite sets of integers}, and we
then speak of \emph{set decision variables} and \emph{set
  constraints}.  We call \emph{universe} the union of the sets in the
domain of a set decision variable; the set value eventually taken by a
decision set variable is then a subset of the universe.

In the following sub-section, we show that constraints are not just a
convenience for high-level modelling, but can also be exploited during
the solution process.

\subsection{Problem Solving by Inference and Search}
\label{sect:solve}

By \emph{intelligent search}, we mean search where at least some form
of inference takes place at every step of the search in order to
reduce the cost of brute-force search \cite{Hooker:integrated}:
\begin{equation*}
  \text{combinatorial problem solving~} =
  \text{~search~} +
  \text{~inference~} + ~\cdots
\end{equation*}
In the following, we discuss two CP ways of solving problems that have
been modelled using constraints.  The classical approach is to perform
systematic search (Section~\ref{sect:systematic}), and a more recent
approach is to trade for time the guarantees of systematic search by
performing stochastic local search (Section~\ref{sect:local}): the two
approaches use completely different forms of inference, which is
encapsulated in reusable fashion within the constraints
(Section~\ref{sect:encaps}).

\subsubsection{Systematic Search}
\label{sect:systematic}

Classically, CP solves combinatorial problems by systematic tree
search, together with backtracking, and performs at every node of the
search tree a particular kind of inference called \emph{propagation}.
For the purpose of this report, we only need to explain the
propagation of a single constraint here: we refer to
\cite{ASTRA:KER:survey} for how to propagate multiple constraints and
how systematic search is conducted.

For each individual constraint, a \emph{propagation algorithm} (or
\emph{propagator}) prunes the domains of its decision variables by
eliminating impossible values according to some desired level of
\emph{consistency}.  For example, under \emph{domain consistency} (DC)
every domain \emph{value} of every decision variable participates in
some solution to the constraint that involves domain values of the
other decision variables.  Also, under \emph{bound consistency} (BC)
every domain \emph{bound} of every decision variable participates in
some solution to the constraint that involves domain values of the
other decision variables.

For example, consider the constraint $\AllDifferent(\Set{x,y,z})$.
Let $x,y$ range over the domain $\Set{1,3}$ and $z$ over
$\Set{1,2,3,4}$: we write $\Domain{x} = \Domain{y} = \Set{1,3}$ and
$\Domain{z} = \Set{1,2,3,4}$ and denote this state by $\Set{x,y
  \mapsto \Set{1,3}, z \mapsto \Set{1,2,3,4}}$.  From this state,
propagation to DC leads to the state $\Set{x,y \mapsto \Set{1,3}, z
  \mapsto \Set{2,4}}$ since $x$ and $y$ must split the values $1$ and
$3$ between themselves so that $z$ cannot take any of these two
values.  From the same start state $\Set{x,y \mapsto \Set{1,3}, z
  \mapsto \Set{1,2,3,4}}$, propagation to BC leads to the state
$\Set{x,y \mapsto \Set{1,3}, z \mapsto \Set{2,3,4}}$ since there do
exist solutions to the constraint where $z$ takes its new lower bound
value~$2$ or its old (and new) upper bound value $4$, so that the
unfeasibility of the intermediate value $3$ is not even checked.  Note
that, in this case, the resulting DC state is strictly stronger than
the resulting BC state: while the initial state encodes a set of $2
\cdot 2 \cdot 4 = 16$ candidate solutions, the BC state encodes a
subset thereof with $2 \cdot 2 \cdot 3 = 12$ candidate solutions, and
the DC state encodes a subset of both with only $2 \cdot 2 \cdot 2 =
8$ candidate solutions, including the $4$ solutions.  From a second
start state $\Set{x,y,z \mapsto \Set{1,2}}$, propagation to DC or BC
leads to the propagator signalling \emph{failure}, because it is
impossible to assign two values to three variables so that the latter
are pairwise distinct.  From a third start state $\Set{x \mapsto
  \Set{1}, y \mapsto \Set{3}, z \mapsto \Set{2,4}}$, propagation to DC
or BC leads to no propagation, but the propagator can simultaneously
detect that all $1 \cdot 1 \cdot 2 = 2$ candidate solutions actually
are solutions, so that the propagator can signal \emph{subsumption}
(or \emph{entailment}).

The propagation of a constraint amounts to \emph{reasoning with
  possible domain values}, but there is \emph{no} obligation to prune
\emph{all} the impossible domain values, as just witnessed when
comparing DC and BC.  If a constraint has multiple propagators
achieving different strengths of consistency (under different time
complexities), then there is a default propagator but the modeller may
also choose one of them, possibly via experiments to find out which
one leads to the best trade-off in search effort; this is the first
non-declarative annotation that may be added to an otherwise
declarative constraint model (and we will encounter a second one
below).  Also, the propagator of a constraint only reasons
\emph{locally}, namely about the decision variables of that
constraint, rather than globally, about all the decision variables of
the entire problem.

\subsubsection{Stochastic Local Search}
\label{sect:local}

Systematic search (as just described) explores the \emph{whole} search
space, though \emph{not} by explicitly trying all possible
combinations of domain values for the decision variables, but
\emph{implicitly} thanks to the interleaving of search with inference.
Suitable values are found \emph{one-by-one} for the decision
variables.  Systematic search offers the guarantee of eventually
finding a solution (or finding and proving an optimal solution, in the
case of an optimisation problem), if one exists, and proving
unsatisfiability otherwise.  However, this may take too long and it
may be more interesting in some situations to find quickly solutions
that may violate some constraints (or may be sub-optimal).  The idea
of \emph{stochastic local search} (SLS; see \cite{Hoos:SLS}, for
example) is to trade this guarantee for speed by not exploring the
whole search space.  Unsatisfiability of the constraints is \emph{a
  priori} not detectable by SLS, and optimality of solutions is a
priori not provable by SLS.

SLS starts from a possibly random assignment of domain values to
\emph{all} the decision variables, without concern for whether some
constraints are violated.  It then tries to find a better assignment
(in the sense of violating fewer constraints, or violating some
constraints less, or yielding a better value of the objective
function) by changing the values of a few decision variables, upon
probing the impacts of many such small changes, which are called
\emph{moves}, and then actually selecting and making one of these
moves.  The set of candidate moves is called the \emph{neighbourhood}.
This iterates, under suitable heuristics and meta-heuristics, until a
sufficiently good assignment has been found, or until some allocated
resource (such as running time or a number of iterations) has been
exhausted.

SLS is an area of intensive research on its own, but the CP concept of
constraint can be usefully imported into SLS, giving rise to what is
known as \emph{constraint-based local search} (CBLS; see \cite{Comet}
for example).  In principle, the declarative part of a constraint
model is thus the \emph{same} as when solving the problem by classical
CP (by systematic tree search interleaved with propagation).  The
inference counterpart of the propagator of a constraint are its
violation functions and its differentiation functions, discussed next.
For the purpose of this report, we only need to explain these
functions for a single constraint here: we refer to
\cite{ASTRA:KER:survey} for how to evaluate them for multiple
constraints and how stochastic local search is conducted.

For each individual constraint, the following functions are required
in a CBLS system:
\begin{itemize}
\item The \emph{constraint violation function} gives a measure of how
  much the constraint is violated under the current assignment.  It
  must be zero if and only if the constraint is satisfied, and
  positive otherwise.
\item The \emph{variable violation function} gives a measure of how
  much a suitable change of a given decision variable may decrease the
  constraint violation.
\item The \emph{assignment delta function} gives the exact or
  approximated increase in constraint violation upon a probed $x
  \IsAssigned d$ assignment move for decision variable $x$ and domain
  value $d$.
\item The \emph{swap delta function} gives the exact or approximated
  increase in constraint violation upon a probed $x:=:y$ swap move
  between two decision variables $x$ and $y$.
\end{itemize}
A constraint or decision variable with higher violation is a stronger
candidate for repair by a move.  A negative delta reflects a decrease
in constraint violation, hence smaller deltas identify better moves.
Differentiation functions for other kinds of moves, such as multiple
assignments, can be added.  Ideally, violations are updated
incrementally in constant time upon the actual making of a move, but
this is not always possible.  Similarly, deltas are ideally computed
differentially in constant time rather than by subtracting the
constraint violations after and before the probed move.

For example, consider the constraint $\AllDifferent(\Set{v,w,x,y,z})$.
Under the assignment $\Set{v \mapsto 4, w \mapsto 4, x \mapsto 5, y
  \mapsto 5, z \mapsto 5}$, the constraint violation could be $3$,
because three variables need to take a suitable new value in order to
satisfy the constraint, and the variable violation of~$y$ could be
$1$, because the constraint violation would decrease by one if $y$
were assigned a suitable new value, such as $6$.  Upon the assignment
moves $y \IsAssigned 4$ and $y \IsAssigned 6$, the constraint
violation increases by $0$ and $-1$, respectively, so the latter
probed move is better.  Upon \emph{any} swap move, the constraint
violation increases by $0$.  When maintaining for every domain value
the number of variables currently taking it, the violations can be
updated in constant time upon an actual move, and the deltas can be
computed in constant time for a probed move.

A neighbourhood can often be designed so that some constraints of the
model remain satisfied if they are satisfied under the starting
assignment.  Such constraints are called \emph{implicit} constraints,
since they need not be given in the constraint model, whereas the
constraints to be satisfied through search are called \emph{explicit}
constraints and must be given in the constraint model.  Since the
explicit constraints can be violated under the current assignment,
they are often called \emph{soft} constraints.  Conversely, since the
implicit constraints can never be violated, they are often called
\emph{hard} constraints.

For example, in a Sudoku puzzle, there is an $\AllDifferent$
constraint on each of the nine rows, columns, and $3 \times 3$ blocks:
the row $\AllDifferent$ constraints can be made implicit upon using a
neighbourhood with swap moves inside rows, since these constraints can
be satisfied under the starting assignment (obtained by generating
nine random permutations of the sequence $\Sequence{1,2,\dots,9}$) and
remain satisfied upon swap moves.

\subsubsection{Conclusion about Search and the Role of Constraints in Search}
\label{sect:encaps}

Both in constraint-based systematic search and in constraint-based
local search, a problem solver software (or simply: \emph{solver})
need only provide the master search algorithm, as well as
implementations of the built-in (meta-)heuristics and constraints that
are used in the problem model.  The modeller is free to design custom
(meta-)heuristics and constraints.  A constraint fully declaratively
encapsulates inference algorithms (propagators or violation functions
and delta functions), which have been written once and for all and are
invoked by the master search algorithm and the (meta-)heuristics in
order to conduct the search for solutions.

The usage of constraints achieves code \emph{reusability}.  It also
entails a clean separation between the declarative and non-declarative
parts of the problem model (which together form the input to the
solver), as well as a clean separation between search and inference
within the solver itself.  The slogan of constraint programming is:
\begin{equation*}
  \text{constraint program~} =
  \text{~model~} +
  \text{~search}
\end{equation*}
because we also have more code \emph{modularity}.

\section{Mathematical Formulation}
\label{sect:math}

We consider two models of airspace sectorisation: one model uses a
decision variable per region, its domain being the set of sectors; the
other model uses a set decision variable per sector, its universe
being the set of regions.  A third approach is to consider both models
at the same time, and add channelling constraints between them that
ensure the two models are consistent.  All work we surveyed
in~\cite{ASTRA:sectorisation:survey} has been done on the first
approach, as the set variable approach might be unrealistic because
the sets could get too large for current CP solvers.  We refer to the
first model as the \emph{graph colouring approach}, and to the second
model as the \emph{set covering approach}.  In both models, there is
the same background data:
\begin{enumerate}
\item A given set $\regions$ of regions, where each region is uniquely
  identified.
\item For each region, there is a given workload value, or, if there
  is more than one type of workload, a tuple of workload values.  The
  set of possible (tuples of) workload values is denoted by
  $\workloaddomain$.  The workload of each region is given by a
  function $\workloadfun \colon \regions \to \workloaddomain$.
\item A function $\workloadcombine$ that takes a set of regions and
  returns the combined workload for that set considered as a sector.
\item Each flight is given as the sequence of regions it visits,
  together with entry and exit time stamps.
\item The background geometry is given as a graph.
\end{enumerate}
In most work, the background geometry was handled implicitly in the
definitions of the constraints, while workload and flight were given
as parameters to the model.

There are three different types of workload: monitoring workload,
conflict workload, and coordination workload.  If all three types of
workload are considered, then the set $\workloaddomain$ is the
Cartesian product $\Nat \times \Nat \times \Nat$, without loss of
generality.  In our previous work \cite{Jaegare:MSc11,ASTRA:ATM13}, we
ignored coordination workload and combined monitoring and conflict
workload into a single integer: the set $\workloaddomain$ was $\Nat$.

In general, the function $\workloadcombine$ performs an addition or a
weighted sum.  It gets complicated when coordination workload is taken
into account, because the coordination between two regions of the same
sector need not be taken into account, so that coordination workload
is not additive.

In what follows, we shall leave the handling of the aggregation of
coordination workload as future work.  We assume that the workload of
a region is given as single integer, and that the workload combination
function $\workloadcombine$ sums up the workloads of the regions in a
region set $R$ constituting a sector, that is:
\[
  \workloadcombine(R) = \sum_{r \in R} \workloadfun(r)
\]

\subsection{Background Geometry}
\label{sec:backgroundgeom}

We assume that the airspace is divided into a countable number of
regions.  These regions can be any shape with flat sides, that is
polytopes.  What is important is that we have a graph connecting
adjacent regions.  In the following, for generality of purpose and
notation, we rename the set $\regions$ into a vertex set $V$.
Formally, the background geometry $G$ induces an undirected graph
\[
  \Tuple{V, E \subseteq V \times V}
\]
where $E$ is a set of undirected edges.  Given $v \in V$, we define
$\Adjacent{v}$ to be the set of vertices that are adjacent to $v$:
\[
  \Adjacent{v} = \Setcomp{w \in V}{\Tuple{v,w} \in E}
\]

When considering constraints on the shape of sectors, we need access
to the shape of its constituent regions.  In particular, if we are
considering regions that are polytopes, then each region has a number
of lower-dimensional facets.  In three dimensions, the facets are
two-dimensional surfaces, and in two dimensions, the facets are
one-dimensional lines.  In algebraic
topology~\cite{May:algebraic_topology}, there is a very general theory
of simplexes, which are representations of geometric objects together
with facets, and facets of facets, all the way down until
zero-dimensional points are reached.  Although we do not need the full
machinery of algebraic topology, the formalisation given here is
inspired by its standard treatment~\cite{May:algebraic_topology}.

Given a graph $G = \Tuple{V, E \subseteq V \times V}$, a \emph{facet
  structure} for $G$ is a set of facets $\facetset$ and a function
$\facetfunname$ from vertices $V$ to facets $\facetset$.  The facet
function needs to satisfy the condition that given any two adjacent
vertices $v$ and $w$ (that is $\Tuple{v,w} \in E$), there is exactly
one facet $f \in \facetset$ such that $f \in \facet{v}$ and $f \in
\facet{w}$.

A facet $f$ is a \emph{border facet} if there is exactly one vertex
$v$ such that $f \in \facet{v}$.  A \emph{border vertex} is a vertex
that has a border facet.

Given a graph $G = \Tuple{V_G, E_G \subseteq V_G \times V_G}$, we
denote the set of border vertices of $G$ by $B_G$.  The
\emph{enveloped graph} of $G$, denoted by $\enveloped{G}$, is the
graph
\[
  \Tuple{
    V_G \Union \Set{\bot},
    E_G \Union (\Set{\bot} \times B_G) \Union (B_G \times \Set{\bot})
  }
\]
That is, there is a unique new vertex $\bot$ that is connected to all
border vertices.  The extended facet function
$\enveloped{\facetfunname}$ is defined as
\[
  \enveloped{\facetfunname}(v) = 
  \begin{cases}
    \facet{v} & \text{if $v \neq \bot$} \\
    \Set{\bot_\text{f}} & \text{otherwise}
  \end{cases}
\]
Given a graph and a facet structure with facet function
$\facetfunname$, the extended facet function satisfies the property
that for any $v \neq \bot$ and $w \neq \bot$ there is \emph{exactly
  one} facet $f \in \facetset$ such that $f \in
\enveloped{\facetfunname}(v)$ and $f \in
\enveloped{\facetfunname}(w)$.

Sometimes, we need to see a vertex set $V$ as an ordered set:
\begin{itemize}
\item Let $v \prec w$ denote that vertex $v$ is to the left of vertex
  $w$ in the vertex set $V$.
\item Let $v \preceq w$ denote that vertex $v$ is to the left of
  vertex $w$ in $V$, with possibly $v=w$.
\item Let $\Pred{v}$ denote the predecessor of vertex $v$ in $V$; if
  $v$ is the leftmost vertex in $V$, then $\Pred{v} = \bot$.
\item Similarly, let $\Succ{v}$ denote the successor of vertex $v$ in
  $V$, if any, else $\bot$.
\end{itemize}
The vertex set $V$ is seen as a sequence rather than as a set whenever
we use the $\prec$ or $\preceq$ relation to specify the semantics of a
constraint.

It will sometimes be useful to have information about the volume of a
region and the surface area of a facet of a region.  We assume that we
are given two functions:
\begin{itemize}
\item $\Volume \colon V \to \Nat$, such that $\Volume(r)$
  returns the volume of region $r$.
\item $\Area \colon V \times \facetset \to \Nat$, such that
  $\Area(r,f)$ returns the area of facet $f$ or region $r$, with $f
  \in \facet{r}$.  Technically, the region $r$ is a redundant
  argument, but we keep it for clarity.
\end{itemize}
We use the terminology of a 3D space.  In a 2D space, the facet area
would be a side length, and the region volume would be a surface area.

\subsection{Flight Data}

We assume without loss of generality that time stamps are given as
integers.  A \emph{flight plan} is a sequence of regions, each with an
entry time stamp and an exit time stamp, such that the times are
increasing between entry and exit time stamps.  Formally a flight plan
$p$ is a member of $(V \times \Nat \times \Nat)^*$ such that if
\[
  p = \Sequence{
    \Tuple{v_1,t_1,t'_1},
    \Tuple{v_2,t_2,t'_2},
    \dots,
    \Tuple{v_m,t_m,t'_m}
  }
\]
then for all $1 \leq i \leq m$ we have $t_i < t'_i$, and for all $1
\leq i < m$ we have $t'_i = t_{i+1}$.  Note that we require a strict
inequality between entry and exit timestamps since aircraft have a
finite velocity.  Let the flight plan of flight $f$ be denoted by
$\Plan(f)$.

\subsection{The Graph Colouring Approach}

Consider a graph $G = \Tuple{V,E}$.  Let the given number of sectors
be $n$.  For each vertex $v$ in $V$, we create a decision variable
$\Col{v}$ with domain $\Set{1,\dots,n}$.  The goal is to assign values
to each decision variable such that we have a valid sectorisation.  A
solution to such a constraint problem is a mapping $\Colour \colon V
\to \Set{1,\dots,n}$.  We assume that solving starts with the special
vertex $\bot$ being given a colour not made available to the vertices
of $G$.

For stochastic local search, the current assignment
$\CurrentAssignment$ induces an undirected graph $\ColourGraph$ with
$V$ as vertex set: there is an edge between vertices $v$ and $w$ if
and only if they are adjacent in $G$ and have the same colour under
$\CurrentAssignment$.  Formally:
\begin{equation}\label{connected:H:cbls}
  \Tuple{v,w} \in \ColourGraph
  \Iff
  \Tuple{v,w} \in E
  \land
  \CurrentValue{\Col{v}} = \CurrentValue{\Col{w}}
\end{equation}
Note that $\ColourGraph$ dynamically changes during search as
$\CurrentAssignment$ changes.

For systematic search, the current state induces an undirected graph
$\ColourGraph$ with $V$ as vertex set: there is an edge between
vertices $v$ and $w$ if and only if they are adjacent in $G$ and
currently have non-disjoint domains, that is they \emph{may}
eventually become part of the same connected component of
$\ColourGraph$.  Formula~(\ref{connected:H:cbls}) becomes:
\begin{equation*}\label{connected:H:cp}
  \Tuple{v,w} \in \ColourGraph
  \Iff
  \Tuple{v,w} \in E
  \land
  \Domain{\Col{v}} \Intersect \Domain{\Col{w}} \neq \EmptySet
\end{equation*}

In a sequence, we call \emph{stretch} a maximal sub-sequence whose
elements are all equal.  For instance, the sequence
$\Sequence{1,1,4,4,4,4,4,4,1,1,1,2}$ has four stretches, namely
$\Sequence{1,1}$, $\Sequence{4,4,4,4,4,4}$, $\Sequence{1,1,1}$, and
$\Sequence{2}$.  A stretch in a sequence is the specialisation for a
one-dimensional geometry of the notion of connected component in
$\ColourGraph$.  In the formalisations of constraints in
Section~\ref{sect:geo}, we use the $\Stretch(X,\ell,r)$ predicate,
which holds if and only if the sequence $X$ has a stretch from index
$\ell$ to index $r$:
\[
    \Stretch(X,\ell,r) \Iff
    X[\Pred{\ell}] \neq X[\ell] = X[r] \neq X[\Succ{r}]
    \land \forall \ell \prec i \prec r : X[i] = X[\ell]
\]
where $X$ is assumed to start and end with a unique value, and $\ell$
and $r$ are neither the first nor the last indices inside $X$.

\subsection{The Set Covering Approach}

Let the given number of sectors be $n$.  We use $n$ set decision
variables $S_1,\dots,S_n$, which are to be assigned subsets of the
universe set $V$.  In addition to the other constraints in the problem
that restrict the assignments of $S_1,\dots,S_n$ to be valid sectors,
we require that all the sets be pairwise disjoint.

\section{Geometric and Workload Constraints}
\label{sect:geo}

We now consider the most important and most representative of the
airspace sectorisation constraints listed in Section~\ref{sect:secto}.
We encapsulate each as a constraint in the constraint programming
sense (as seen in Section~\ref{sect:model}), give its semantics, and
design the encapsulated inference algorithms, namely propagators for
systematic search (as seen in Section~\ref{sect:systematic}), as well
as violation and differentiation functions for stochastic local search
(as seen in Section~\ref{sect:local}).  It is important to keep in
mind that these algorithms and functions are designed in a
problem-\emph{independent} fashion, so as to make the constraints
highly reusable.

We leave the set colouring approach as future work, and assume the
graph colouring approach for all the constraints:
\begin{itemize}
\item We consider a graph $G$ with vertex set $V$ and edge set $E$.
\item We look for a mapping $\Colour \colon V \to \Colours$, that is
  we create a colour decision variable $\Col{v}$ for every vertex $v
  \in V$.  Let $\Colours = \bigcup_{v \in V} \Domain{\Col{v}}$ be the
  set of available colours.
\item Let $\Value$ be a sequence, indexed by $V$, of integer values.
\end{itemize}
The notation $\HigherOrder{\phi}$ uses the Iverson bracket (typeset in
\red{red} for the convenience of those viewing this document in colour) to
represent the truth of formula $\phi$, with truth represented by $1$
and falsity by $0$.

\subsection{Connectedness}
\label{sect:connected}

The $\Connected(G,\Colour,\RelOp,N)$ constraint, with $\RelOp \in
\Set{\leq,<,=,\neq,>,\geq}$, holds if and only if the number of
colours used in the sequence $\Colour$ is in relation $\RelOp$ with
$N$, and there is a path in the graph $G = \Tuple{V,E}$ between any
two vertices of the same colour that only visits vertices of that
colour.  Formally:
\begin{equation}\label{connected:sem1}
  \begin{array}{c}
    \Cardinality{\Setcomp{\Col{v}}{v \in V}} ~\RelOp~ N
    ~\land \\
    \forall v,w \in V :
    \Col{v} = \Col{w}
    \Implies
    \Tuple{v,w} \in E^*_{\Col{v}}
  \end{array}
\end{equation}
where $E^*_c$ denotes the transitive closure of $E_c$, which
is the adjacency relation $E$ projected onto adjacency for
vertices of colour $c$:
\[
  E_c = \Setcomp{\Tuple{v,w} \in E}{\Col{v} = c = \Col{w}}
\]
Hence the total number of connected components in the induced graph
$\ColourGraph$ must be in relation $\RelOp$ with $N$, and the number
of connected components per colour must be at most $1$.

\paragraph{Arbitrary Number of Dimensions.}
The $\Connected(G,\Colour,\RelOp,N)$ constraint generalises the main
aspect of the $\Constraint{Connect\_Points}(w,h,d,\Colour,N)$
constraint of~\cite{GC-catalogue}, which considers $\RelOp$ is ``$=$''
and considers $G$ to be induced by a $w \times h \times d$ cuboid
divided into same-sized regions; further, there is a special colour
(value $0$) for which there is no restriction on the number of
connected components.  An initial idea for a propagator is outlined
in~\cite{GC-catalogue}: we flesh it out below and also give violation
and differentiation functions.

The constraint $\Connected$ is a hard constraint in our prior work on
airspace sectorisation using stochastic local
search~\cite{ASTRA:ATM13}, hence no violation and differentiation
functions are given there.  A connectedness constraint was
accidentally forgotten in our prior work on airspace sectorisation
using systematic search~\cite{Jaegare:MSc11}, hence no propagator is
given there.

\paragraph{One Dimension.}
The constraint $\Connected(G,\Colour,\RelOp,N)$ for a graph $G$
induced by a one-dimensional geometry (in which connected components
are called \emph{stretches}) generalises the main aspect of the
$\Constraint{Multi\_Global\_Contiguity}(\Colour)$
constraint~\cite{GC-catalogue}, which itself generalises the
$\Constraint{Global\_Contiguity}(\Colour)$ constraint
of~\cite{Maher:contiguity}: the latter constrains only one colour
(value~$1$) but the former constrains several colours, and both lack
the decision variable $N$ and hence $\RelOp$; further, both have a
special colour (value~$0$) for which there is no restriction on the
number of stretches (there are at most two such stretches when there
is only one constrained colour).  A propagator for
$\Constraint{Global\_Contiguity}$ is given in~\cite{Maher:contiguity},
and a decomposition based on the $\Automaton$ constraint is given
in~\cite{GC-catalogue}: we generalise these ideas below and also give
violation and differentiation functions.

The trajectory-based convexity of an airspace sectorisation is
achieved by posting for every flight a $\Connected(G,\Colour,\leq,s)$
constraint on the sequence $\Colour$ of decision variables denoting
the sequence of colours of its \emph{one}-dimensional visited region
sequence $V$, where $s$ is the imposed or maximum number of sectors.
The initial domain of each colour decision variable $\Col{v}$ is
$\Set{1,2,\dots,s}$.

Such a trajectory-based convexity constraint is called the
$\Constraint{Contiguity}$ constraint in our prior work on airspace
sectorisation under systematic search~\cite{Jaegare:MSc11}, but lacks
the decision variable $N$ and hence $\RelOp$; further, the propagator
outlined there prunes less strongly than the one describe below (as it
lacks both steps~2).

Such a trajectory-based convexity constraint is a soft constraint in
our prior work on airspace sectorisation under stochastic local
search~\cite{ASTRA:ATM13}, but lacks the decision variable $N$ and
hence $\RelOp$; further, the constraint violation is defined
differently there (in a manner that requires an asymptotically higher
runtime to compute than the one we give below), and the variable
violation and differentiation functions are not given there (though
they are in the unpublished code underlying the experiments).

\subsubsection{Violation and Differentiation Functions}
\label{sect:connected:cbls}

The violation and differentiation functions described below have no
asymptotically better specialisation to the case of $G$ being induced
by a one-dimensional space.  Hence they apply to both connectedness in
a space of an arbitrary number of dimensions and to contiguity in a
one-dimensional space.

\paragraph{Soft Constraint.}
If the $\Connected$ constraint is considered explicitly, then we
proceed as follows.  For representing the induced graph
$\ColourGraph$, we show that it suffices to initialise and maintain
the following two data structures, which are internal to the
constraint:
\begin{itemize}
\item Let $\NCCcolour{c}$ denote the number of connected components
  (CCs) of $\ColourGraph$ whose vertices currently have colour $c$.
\item Let $\NCC$ denote the current number of connected components of
  $\ColourGraph$:
  \begin{equation}\label{connected:NCC:cbls}
    \NCC = \sum_{c \in \Colours} \NCCcolour{c}
  \end{equation}
\end{itemize}
We can now re-formalise the semantics~(\ref{connected:sem1}): the
$\Connected(G,\Colour,\RelOp,N)$ constraint is satisfied if and only
if
\begin{equation}\label{connected:sem2}
  \NCC ~\RelOp~ N
  \land
  \forall c \in Colours : \NCCcolour{c} \leq 1  
\end{equation}

The \emph{violation of a colour decision variable}, say $\Col{v}$ for
vertex $v$, is the current excess number, if any, of connected
components of $\ColourGraph$ for the colour of $v$:
\begin{equation}\label{connected:colour:vio}
  \VarViolation{\Col{v}} = \NCCcolour{\CurrentValue{\Col{v}}}-1
\end{equation}
This variable violation is zero if $v$ currently has a colour for
which there is exactly one connected component in $\ColourGraph$, and
positive otherwise.

The \emph{violation of the counter decision variable} $N$ is $0$ or
$1$ depending on whether $\NCC$ is in relation $\RelOp$ with the
current value of $N$:
\begin{equation}\label{connected:counter:vio}
  \VarViolation{N} =
  1 - \HigherOrder{\NCC ~\RelOp~ \CurrentValue{N}}
\end{equation}
This variable violation is zero if $\NCC ~\RelOp~ \CurrentValue{N}$
currently holds, and one otherwise.

The \emph{violation of the constraint} is the sum of the variable
violation of $N$ and the current excess number, if any, of connected
components for all colours:
\begin{equation}\label{connected:cons:vio}
  \Violation
  = \VarViolation{N}
  + \sum_{c \in \Colours} \max(\NCCcolour{c}-1,~0)
\end{equation}
The constraint violation is zero if $\NCC ~\RelOp~ \CurrentValue{N}$
holds and there currently is at most one connected component in
$\ColourGraph$ for each colour.

The impact on the constraint violation of a colour assignment move
$\Col{v} \IsAssigned c$ is measured by the following \emph{colour
  assignment delta} function:
\begin{equation}\label{connected:flip}
  \begin{array}{l}
    \Probe{\Col{v} \IsAssigned c} = \\
    \LetIn{
      p = \HigherOrder{\forall w \in \Adjacent{v} :
        \CurrentValue{\Col{w}} \neq c} \\
      ~~~~m = \HigherOrder{\forall w \in \Adjacent{v} :
        \CurrentValue{\Col{w}} \neq \CurrentValue{\Col{v}}}
    }
    {
      p
      - m
      + \HigherOrder{\NCC     ~\RelOp~ \CurrentValue{N}}
      - \HigherOrder{\NCC+p-m ~\RelOp~ \CurrentValue{N}}
    }
  \end{array}
\end{equation}
Indeed, only the violation of the unchanged $N$ and the numbers of
connected components of the old and new colour of vertex $v$ need to
be considered.
On the one hand, the number of connected components of colour $c$
increases by one if all vertices adjacent to $v$ do not currently have
its new colour $c$, so that $v$ forms a new connected component of
colour~$c$.
On the other hand, the number of connected components of colour
$\CurrentValue{\Col{v}}$ decreases by one if all vertices adjacent to
$v$ do not currently have that original colour of~$v$, so that~$v$ was
the last element of some connected component of colour
$\CurrentValue{\Col{v}}$.
To avoid increasing the number of connected components and increase
the likelihood of decreasing their number, it is advisable to use a
neighbourhood where vertices at the border of a connected component
are re-coloured using a currently unused colour or the colour of an
adjacent connected component.
An assignment move on vertex $v$ can be differentially probed in time
linear in the degree of $v$ in $G$.

A colour swap move $\Col{v} :=: \Col{w}$, where vertices $v$ and $w$
exchange their colours, is the sequential composition of the two
colour assignment moves $\Col{v} \IsAssigned \CurrentValue{\Col{w}}$
and $\Col{w} \IsAssigned \CurrentValue{\Col{v}}$.  The \emph{colour
  swap delta} is the sum of the deltas for these two moves (upon
incrementally making the first move), and there is no asymptotically
faster way to compute this delta, as the complexity of probing an
assignment move does not depend on the number of vertices.

The additive impact on the constraint violation of a counter
assignment move $N \IsAssigned n$ is measured by the following
\emph{counter assignment delta} function:
\begin{equation*}
  \Probe{N \IsAssigned n}
  = \HigherOrder{\NCC ~\RelOp~ \CurrentValue{N}}
  - \HigherOrder{\NCC ~\RelOp~ n}
\end{equation*}
The violation increases by one if the number of connected components
was but now is not in relation $\RelOp$ with $N$.  It decreases by one
if the number of connected components now is but was not in relation
$\RelOp$ with $N$.  An assignment move on $N$ can be differentially
probed in constant time.

To achieve \emph{incrementality}, once a move has been picked and
made, the internal data structures and the variable and constraint
violations must be updated.  From the colour assignment delta
function~(\ref{connected:flip}), the following updating code for a
colour assignment move $\Col{v} \IsAssigned c$ follows directly, where
the new assignment $\NewAssignment$ is the old assignment
$\CurrentAssignment$, except that $\NewValue{\Col{v}} = c$:
\begin{algorithmic}[1]
  \IF[$v$ forms a new CC of colour $c$]
     {$\forall w \in \Adjacent{v} :
       \CurrentValue{\Col{w}} \neq c$}
    \STATE $\NCCcolour{c} \IsAssigned \NCCcolour{c} + 1$
    \STATE $\NCC \IsAssigned \NCC + 1$
    \STATE $\Violation \IsAssigned \Violation + 1$
  \ENDIF
  \IF[$v$ formed a CC of $\CurrentValue{\Col{v}}$]
     {$\forall w \in \Adjacent{v} :
       \CurrentValue{\Col{w}} \neq \CurrentValue{\Col{v}}$}
    \STATE $\NCCcolour{\CurrentValue{\Col{v}}}
            \IsAssigned \NCCcolour{\CurrentValue{\Col{v}}} - 1$
    \STATE $\NCC \IsAssigned \NCC - 1$
    \STATE $\Violation \IsAssigned \Violation - 1$
  \ENDIF
  \FORALL{$v \in V$}
    \STATE $\VarViolation{\Col{v}} \IsAssigned \NewValue{\NCCcolour{\Col{v}}}-1$
  \ENDFOR
  \STATE $\VarViolation{N} \IsAssigned 1 - \HigherOrder{\NCC ~\RelOp~ \CurrentValue{N}}$
\end{algorithmic}
Incremental updating for a colour assignment move takes time linear in
the degree of vertex $v$ in $G$ and linear in the number
$\Cardinality{V}$ of vertices (and colour decision variables).  Code
follows similarly for the colour swap and counter assignment moves.

For the remaining constraints, we assume that some relationships among
internal data structures, such as~(\ref{connected:NCC:cbls}), and the
violation functions, such as~(\ref{connected:colour:vio})
to~(\ref{connected:cons:vio}), are defined as
\emph{invariants}~\cite{Comet}, so that the solver automatically
updates these quantities incrementally, without the constraint
designer having to write explicit code, such as lines~3, 4, and~7
to~11.

\paragraph{Hard Constraint.}
If the $\Connected$ constraint is considered implicitly, as in
our~\cite{ASTRA:ATM13}, then it can be satisfied cheaply in the start
assignment, by partitioning $G$ into connected components and setting
$N$ according to $\RelOp$, and maintained as satisfied upon every
move, by only considering moves that re-colour a vertex at the border
of a connected component to the colour of an adjacent connected
component.  For instance, if $\RelOp$ is equality, then one can
partition $G$ into $n = \max(\Domain{N})$ connected components and set
$N \IsAssigned n$.

\subsubsection{Propagator}
\label{sect:connected:cp}

\paragraph{One Dimension.}
If $G$ is induced by a one-dimensional space, then
propagation goes as follows.
 
If a vertex $v$ is given a colour $c$ (by obtaining $\Domain{\Col{v}}
= \Set{c}$ in the current state through propagation of either another
constraint or a search decision), then:
\begin{itemize}
\item Let $\ell$ be the rightmost vertex, if any, \emph{to the left}
  of $v$ where $c \notin \Domain{\Col{\ell}}$.
\item Let $r$ be the leftmost vertex, if any, \emph{to the right} of
  $v$ where $c \notin \Domain{\Col{r}}$.
\item Let $\ell'$ be the rightmost vertex with $\ell' \preceq v$ where
  $\Domain{\Col{\ell'}} = \Set{c}$.
\item Let $r'$ be the leftmost vertex with $v \preceq r'$ where
  $\Domain{\Col{r'}} = \Set{c}$.
\end{itemize}
All occurrences of colour $c$ must be strictly between vertices $\ell$
and $r$.  All vertices from $\ell'$ to $r'$ must get colour $c$.  Hence
we prune as follows:
\begin{itemize}
\item If $\ell$ exists, then:
  \begin{enumerate}
  \item Prune value $c$ from the domain of every vertex to the left of
    $\ell$.
  \item If $\Domain{\Col{\ell}} = \Set{d}$, then prune value $d$
    from the domain of every vertex to the right of $v$.
  \end{enumerate}
\item If $r$ exists, then:
  \begin{enumerate}
  \item Prune value $c$ from the domain of every vertex to the right
    of $r$.
  \item If $\Domain{\Col{r}} = \Set{d}$, then prune value $d$ from
    the domain of every vertex to the left of $v$.
  \end{enumerate}
\item Set $\Domain{\Col{i}} = \Set{c}$ for every vertex $i$ with
  $\ell' \prec i \prec r'$.
\end{itemize}
To propagate on $N$, it is best for the propagator to have an internal
data structure: see the multi-dimensional case for details.

This propagator is only worth invoking when the domain of one of its
decision variables shrinks to a singleton.  It achieves domain
consistency.  No propagation is possible when the domain of a decision
variable loses a value without becoming a singleton.

\paragraph{Arbitrary Number of Dimensions.}
If $G$ is induced by a space of more than one dimension, then the
notions of `left' and `right' have to be generalised.  We assume that
the propagator initialises (at its first invocation) and updates (at
subsequent invocations) the following internal data structures, which
are strongly related to those we use in
Section~\ref{sect:connected:cbls} for stochastic local
search:\footnote{The strong relationship between internal
  datastructures needed for the violation and differentiation
  functions of stochastic local search and internal data structures
  needed for propagators of systematic search was not obvious to the
  authors at the outset of this research: this issue is worth
  investigating further, as code generation might be possible.}
\begin{itemize}
\item The induced graph $\ColourGraph$ (which we define but do not
  store for stochastic local search).
\item Let $\NCCcolour{c}$ denote the number of connected components of
  $\ColourGraph$ whose vertices currently have colour $c$ in their
  domains, with $c \in \Colours$.
\item Let $\NCC$ denote the current sum of all $\NCCcolour{c}$, as in
  formula~(\ref{connected:NCC:cbls}).
\end{itemize}
The semantics~(\ref{connected:sem2}) of the constraint gives us a
constraint checker and a feasibility test.

Upon pruning of values (by propagation of either another constraint or
a search decision) from the domain of a colour decision variable
$\Col{v}$, we (possibly incrementally) update the internal data
structures and use the graph invariants
of~\cite{Beldiceanu:graphProps:bounds} in order to obtain possible
pruning on other colour decision variables if not the counter decision
variable $N$.  We conjecture that this achieves domain consistency,
like in the one-dimensional case.

Upon pruning of values from the domain of the counter decision
variable $N$, we presently do not know whether pruning is possible on
the colour decision variables.

\paragraph{Discussion.}
For many choices of $\RelOp$ and $N$, the $\Connected$ constraint is
easy to satisfy, namely by colouring all vertices with the same
colour, so that there is only one connected component.  In other
words, the density of solutions to the $\Connected$ constraint may be
very high within the Cartesian product of the domains of its decision
variables.  Such a situation is not very conducive to propagation, as
discussed next.

Starting from full colour domains for all vertices, there potentially
is only one connected component until the underlying space is cut into
at least two sub-spaces by colouring an entire swathe of vertices in
some colour that has already been eliminated for other vertices.  Only
a very specific search procedure would achieve this only situation
that is conducive to domain pruning.  While this situation often
arises naturally in a one-dimensional space, it does not do so in a
multi-dimensional space and systematic search with interleaved
propagation essentially degenerates into generate-and-test search,
because the propagator of $\Connected$ can only be invoked near the
leaves of the search tree.

In conclusion, we are pessimistic about the utility of a propagator
for the $\Connected$ constraint in a multi-dimensional space, at least
when following the graph colouring approach.  Unless a representation
more conducive to propagation can be found, we advocate stochastic
local search over systematic search in the presence of this
constraint.

\subsection{Compactness}
\label{sect:compact}

Consider a graph $G = \Tuple{V,E}$ induced by a space of at least two
dimensions.  The $\Compact(G,\Colour,t)$ constraint holds if and only
if the sum of the sphericity discrepancies of the connected components
of the graph $\ColourGraph$ induced by $G$ and the sequence $\Colour$
is at most the threshold $t$, whose value is given.

The sphericity discrepancy of a connected component is defined as
follows.  Recall that in $G$ each facet is endowed with a surface
area, and each vertex is endowed with a volume.  We define the
following concepts:
\begin{itemize}
\item In Section~\ref{sec:backgroundgeom}, we formalised the notion of
  border vertices, that is vertices at the edge of the geometry.  Here
  we are dealing with colouring, so we need to formalise the border of
  coloured regions.  A facet $f$ of a vertex $v$ is a \emph{border
    facet under a sequence} $\Colour$ if $v$ has no adjacent vertex
  for $f$ or $v$ has a different colour than the adjacent vertex that
  shares $f$.  Formally, a facet $f \in \facet{v}$ of a vertex $v$ is
  a border facet if and only if the following statement
  \[
    \Col{w} \neq \Col{v}
  \]
  holds for \emph{the} vertex $w$ that shares $f$ with $v$.

  To simplify the formalisation, we assume that we are working with
  background geometry of the form $\enveloped{G}$ for some given
  background geometry $G$.  That is, there is a unique special vertex
  in $V$, called $\bot$, that shares a facet with every vertex where
  there otherwise is no adjacent vertex for that facet.  Now a vertex
  has exactly one adjacent vertex for each of its facets.
\item The \emph{border surface area} of a set $W$ of vertices that
  have the same colour under a sequence $\Colour$ (such as a connected
  component of the induced graph $\ColourGraph$), denoted by $A_W$, is
  the sum of the surface areas of the border facets of the vertices in
  $W$:
  \begin{equation}\label{def:borderSurfaceArea}
    A_W =
    \sum_{v \in W}
    \sum_{\substack{
        w \in \Adjacent{v} \\
        f \in \facet{v} \Intersect \facet{w} \\
        w \notin W}
    }
    \HigherOrder{\Col{w} \neq \Col{v}} \cdot \Area(v,f)
  \end{equation}
  Recall that $\Cardinality{\facet{v} \Intersect \facet{w}} = 1$ when
  vertices $v$ and $w$ are adjacent.
\item The \emph{sphere surface area} of a set $W$ of vertices that
  have the same colour under a sequence $\Colour$, denoted by $S_W$,
  is the surface area of a sphere that has as volume the total volume
  $V_W$ of the vertices in $W$ (see~\cite{sphericity} for the
  derivation of this formula):
  \[
    S_W = \pi^{1/3} \cdot (6 \cdot V_W)^{2/3}
  \]
  where
  \[
    V_W = \sum_{w \in W} \Vol{w}
  \]
  In case $G$ is induced by a 3D cuboid divided into same-sized
  regions, we can rather define the sphere surface area as the surface
  area of the smallest collection of regions that contains the sphere
  that has as volume the total volume of the vertices in $W$.  We omit
  the mathematical details, as they are specific to the shape of the
  regions: a space can be tiled by any kind of polyhedra, such as
  cubes or beehive cells.
\item The \emph{sphericity discrepancy} of a set $W$ of vertices that
  have the same colour under a sequence $\Colour$, denoted by
  $\Psi_W$, is the difference between the border surface area of $W$
  under $\Colour$ and the sphere surface area of $W$ under $\Colour$:
  \[
    \delta\Psi_W = A_W - S_W
  \]
  We derive this concept from the \emph{sphericity} $\Psi$ of a shape
  $p$, defined in~\cite{sphericity} to be the ratio between the
  surface area $S_p$ of a sphere that has the same volume as $p$ and
  the surface area $A_p$ of $p$; note that $\Psi=1$ if $p$ is a
  sphere, and $0<\Psi<1$ otherwise, assuming $p$ is not empty.  Our
  concept would be defined as the subtraction $A_p-S_p$ rather than as
  the ratio $S_p/A_p$, as we need (for stochastic local search) a
  non-negative metric that is $0$ in the good case, namely when $p$ is
  (the smallest over-approximation of) a sphere.
\end{itemize}

Note that the $\Compact$ constraint imposes no limit on the number of
connected components of $\ColourGraph$ per colour: if there are
several connected components for a colour, then the total sphericity
discrepancy may be unnecessarily large.

Such a compactness constraint is a soft constraint in our prior work
on airspace sectorisation under stochastic local
search~\cite{ASTRA:ATM13}, but lacks the threshold $t$ there; we
generalise the ideas of its violation and differentiation functions
using the concept of sphericity discrepancy, and we describe them in
much more detail.

There was no compactness constraint in our prior work on airspace
sectorisation using systematic search~\cite{Jaegare:MSc11}, and we are
not aware of any published propagator for any such constraint.

\subsubsection{Violation and Differentiation Functions}
\label{sect:compact:cbls}

\paragraph{Soft Constraint.}
If the $\Compact$ constraint is considered explicitly, then we proceed
as follows.  We initialise and incrementally maintain the following
data structures, which are internal to the constraint:
\begin{itemize}
\item Let $\Border{v}$ denote the border surface area of the vertex
  set $\Set{v}$.  If every vertex only has facets of unit surface area
  (for instance, when $G$ is induced by a space divided into
  same-sized cubes or squares), then $\Border{v}$ is defined as
  follows:
  \begin{equation*}
    \Border{v} =
    \sum_{w \in \Adjacent{v}}
    \HigherOrder{\CurrentValue{\Col{w}} \neq \CurrentValue{\Col{v}}}
  \end{equation*}
  Otherwise, the formula needs to be generalised as follows,
  using~(\ref{def:borderSurfaceArea}):
  \begin{equation*}
    \Border{v} =
    \sum_{\substack{w \in \Adjacent{v} \\ f \in \facet{v} \Intersect \facet{w}}}
    \HigherOrder{\CurrentValue{\Col{w}} \neq \CurrentValue{\Col{v}}}
    \cdot \Area(v,f)
  \end{equation*}
\item Let $\CCs$ denote the current set of connected components of the
  induced graph $\ColourGraph$, each encoded by a tuple
  $\Tuple{\sigma,\nu}$, meaning that it currently has total surface
  area $\sigma$ and total volume $\nu$.
\end{itemize}

The \emph{violation of a decision variable}, say $\Col{v}$ for vertex
$v$, is its current weighted border surface area:
\begin{equation*}
  \VarViolation{\Col{v}} = f(\Border{v})
\end{equation*}
where the weight function $f$ can be the identity function, but can
also suitably penalise larger border surface areas, provided $f(0)=0$;
in~\cite{ASTRA:ATM13}, we found that using $f$ as $\lambda x : x^2$
works well enough.  The variable violation is zero if $v$ is not a
border facet.

The \emph{violation of the constraint} is the current excess, if any,
of the sum of the sphericity discrepancies of the connected
components:
\begin{equation}\label{compact:cons:vio}
  \Violation =
  \max\left(
    \sum_{\Tuple{\sigma,\nu} \in \CCs}
    \left(\sigma - \pi^{1/3} \cdot (6 \cdot \nu)^{2/3}\right) - t,~ 0
  \right)
\end{equation}
The constraint violation is zero if the total sphericity discrepancy
does not exceed $t$.

To define the additive impact $\Probe{\Col{v} \IsAssigned c}$ on the
constraint violation of an assignment move $\Col{v} \IsAssigned c$, we
first define the impact on the violation of a variable $\Col{w}$ for a
vertex $w$ that shares a facet $f$ with $v$:
\begin{equation*}
  \begin{array}{l}
    \ProbeVar{\Col{w}}{\Col{v} \IsAssigned c} \\
    =
    \begin{cases}
      - \Area(v,f) &
      \text{if~~} \CurrentValue{\Col{v}} \neq \CurrentValue{\Col{w}} = c
      \land w \in \Adjacent{v} 
      \land f \in \facet{v} \Intersect \facet{w} \\
      + 0 &
      \text{if~~} \CurrentValue{\Col{v}} \neq \CurrentValue{\Col{w}} \neq c \\
      & \hspace{.8mm}
      \lor~ \CurrentValue{\Col{v}} = \CurrentValue{\Col{w}} = c \\
      + \Area(v,f) &
      \text{if~~} \CurrentValue{\Col{v}} = \CurrentValue{\Col{w}} \neq c
      \land w \in \Adjacent{v} 
      \land f \in \facet{v} \Intersect \facet{w} \\
    \end{cases}
  \end{array}
\end{equation*}
This differential probing can be done in constant time.

The impact on the violation of $\Col{v}$ itself is the sum of the
impacts on the violations of the variables corresponding to the
vertices adjacent to $v$:
\begin{equation}\label{compact:sum:flip}
  \ProbeVar{\Col{v}}{\Col{v} \IsAssigned c} =
  \sum_{w \in \Adjacent{v}} \ProbeVar{\Col{w}}{\Col{v} \IsAssigned c}
\end{equation}
This differential probing can be done in time linear in the degree of
$v$ in $G$.

The impact $\Probe{\Col{v} \IsAssigned c}$ on the constraint violation
of an assignment move $\Col{v} \IsAssigned c$ can unfortunately not be
computed cheaply.  The connected component containing $v$ with its
\emph{original} colour $\CurrentValue{\Col{v}}$ may be split into
several components after the move.  Further, the connected component
containing $v$ with its \emph{new} colour $c$ may be the merger of
several components existing prior to the move.  The same holds for the
connected components containing the vertices adjacent to $v$: some of
them may merge after the move.  Hence the only way to measure the
\emph{exact} impact using all the connected components is first to
compute the new connected components, in $\Theta(V+E)$ time, as well
as their surface area and volume attributes, and then to subtract the
old violation from the new violation computed
using~(\ref{compact:cons:vio}).  One may of course choose not to probe
at all the impact on the violation of the $\Compact$ constraint during
search.  One may also be content with a much more cheaply computed
\emph{approximate} impact, namely $\ProbeVar{\Col{v}}{\Col{v}
  \IsAssigned c}$, which as we saw can be computed in time linear in
the degree of $v$ in $G$.

If probing \emph{is} used during search, the described exact probing
in time linear in the size of $G$ is considered too expensive, and the
described fast approximate probing is deemed too risky (as it can be
an under-approximation, which is dangerous~\cite{ASTRA:JoH}), then we
propose another measure of constraint violation, which does not need
the data structure $\CCs$ but leads to a coarser approximation of
sphericity.  This requires first changing the semantics of the
constraint: the $\Compact(G,\Colour,t)$ constraint holds if and only
if the sum of the border surface areas of all vertices is at most the
threshold $t$, whose value is given.  We then redefine the
\emph{violation of the constraint} to be the excess, if any, of the
current total weighted border surface area of all vertices:
\begin{equation}\label{compact:cons:vio2}
  \Violation =
  \max\left(
   \frac{1}{2} \cdot \sum_{i \in V} \VarViolation{\Col{i}} - t,~ 0
  \right)
\end{equation}
The factor $\frac{1}{2}$ compensates for every shared facet being
counted twice.  The constraint violation is zero if the current total
weighted border surface area does not exceed $t$.

The impact $\Probe{\Col{v} \IsAssigned c}$ on the constraint violation
of an assignment move $\Col{v} \IsAssigned c$ can now be computed
cheaply.  Let $s$ be the current value of the expression $\frac{1}{2}
\cdot \sum_{i \in V} \VarViolation{\Col{i}} - t$
inside~(\ref{compact:cons:vio2}).  Since the violations of the
variables corresponding to $v$ and its adjacent vertices are the only
terms that potentially change in the evaluation of $s$, that
expression changes due to~(\ref{compact:sum:flip}) by
$\ProbeVar{\Col{v}}{\Col{v} \IsAssigned c}$ only.  We get the
following \emph{assignment delta} function:
\begin{equation*}
  \Probe{\Col{v} \IsAssigned c} =
  \max(s + \ProbeVar{\Col{v}}{\Col{v} \IsAssigned c},~ 0)
  - \max(s,~ 0)
\end{equation*}
An assignment move on vertex $v$ can be differentially probed in time
linear in the degree of $v$ in $G$.

It is advisable to use a neighbourhood where vertices at the border of
a connected component are re-coloured using a currently unused colour
or the colour of an adjacent connected component.

A swap move $\Col{v} :=: \Col{w}$, where vertices $v$ and $w$ exchange
their colours, is the sequential composition of the two assignment
moves $\Col{v} \IsAssigned \CurrentValue{\Col{w}}$ and $\Col{w}
\IsAssigned \CurrentValue{\Col{v}}$.  The \emph{swap delta} is the sum
of the assignment deltas for these two moves (upon incrementally
making the first move), and there is no asymptotically faster way to
compute this delta, as the complexity of probing an assignment move
does not depend on the number of vertices.

We omit the rather clerical code for achieving incrementality.

\paragraph{Hard Constraint.}
It is very difficult to consider the $\Compact$ constraint implicitly,
as it is not obvious how to satisfy it cheaply in the start assignment
and under what probed moves to maintain it satisfied.

\subsubsection{Propagator}
\label{sect:compact:cp}

Even more so than a propagator for the $\Connected$ constraint (see
Section~\ref{sect:connected:cp}), a propagator for the $\Compact$
constraint will not be able to perform much domain pruning until most
of the decision variables have singleton domains.  Indeed, there is no
useful particular case of $\Compact$ for one-dimensional spaces
(unlike for $\Connected$, where in a one-dimensional space
connectedness reduces to stretch contiguity, for which significant
pruning is possible, even to domain consistency), and for $\Compact$
there can be several connected components per colour (unlike for
$\Connected$, where every colour has at most one connected component).
In fact, if the threshold $t$ is not exceeded by the sphericity
discrepancy of the entire underlying space, then the $\Compact$
constraint is easy to satisfy (like $\Connected$), namely by colouring
all vertices with the same colour, so that there is only one connected
component.  For the implications of this insight, see the discussion
of propagation of the $\Connected$ constraint in
Section~\ref{sect:connected:cp}.

\subsection{Minimum Dwell Time: Minimum Stretch Sum}
\label{sect:stretchSum}

Consider a graph $G = \Tuple{V,E}$ induced by a one-dimensional space,
so that the two sequences $\Colour$ and $\Value$ are indexed by a
vertex sequence $V$ rather than vertex set.  The
$\StretchSum(G,\Colour,\Value,\RelOp,t)$ constraint, with $\RelOp \in
\Set{\leq,<,=,\neq,>,\geq}$, holds if and only if every stretch of the
sequence $\Colour$ corresponds to a subsequence of $\Value$ whose sum
is in relation $\RelOp$ with threshold $t$, whose value is given.
Formally:
\begin{equation*}
  \begin{array}{c}
    \forall \ell \preceq r \in V :
    \Stretch(\Colour,\ell,r) \\
    \Implies
    \left( \displaystyle\sum_{\ell \preceq v \preceq r} \Val{v} \right)
    ~\RelOp~ t
  \end{array}
\end{equation*}
Note that there is no limit on the number of stretches per colour.

In airspace sectorisation, every flight entering a sector must stay
within it for a given minimum amount of time (say $t=120$ seconds), so
that the coordination work pays off and that conflict management is
possible.  This minimum dwell-time constraint is achieved by posting
for every flight $f$ a $\StretchSum(G,\Colour,\Value,\geq,120)$
constraint on the sequence $\Colour$ of decision variables denoting
the sequence of colours of its visited region sequence~$V$, with
$\Value$ storing the durations of the flight $f$ in each region:
\[
  \Value = \Sequence{t'_i - t_i \mid \Tuple{\_,t_i,t'_i} \in \Plan(f)}
\]

The $\StretchSum$ constraint is a soft constraint in our prior work on
airspace sectorisation under stochastic local
search~\cite{ASTRA:ATM13}, but the constraint violation is defined
differently there (in a manner that requires an asymptotically higher
runtime to compute than the one we give below), and the variable
violation and differentiation functions are not given there (though
they are in the unpublished code underlying the experiments).

The $\StretchSum$ constraint is called the $\Constraint{SlidingSum}$
constraint in our prior work on airspace sectorisation under
systematic search~\cite{Jaegare:MSc11}, but the propagator outlined
there is very different from the one we describe below, as it is only
worth invoking when the domain of one of its decision variables
shrinks to a singleton.

\subsubsection{Violation and Differentiation Functions}
\label{sect:stretchSum:cbls}

\paragraph{Soft Constraint.}
If the $\StretchSum$ constraint is considered explicitly, then we
proceed as follows.  For simplicity of notation, we assume that
$\RelOp$ is $\geq$.  The other values of $\RelOp$ are handled
analogously.  We initialise and incrementally maintain the following
data structure, which is internal to the constraint:
\begin{itemize}
\item Let $\StretchVertex{v}$ denote the tuple
  $\Tuple{\ell,r,c,\sigma}$, meaning that vertex $v \in V$ is
  currently in a colour stretch, from vertex $\ell$ to vertex $r$,
  whose colour is $c$ and value sum is $\sigma$:
  \[
    \sigma = \sum_{\ell \preceq i \preceq r} \Val{i}
  \]
\end{itemize}
We say that a colour stretch with value sum $\sigma$ is a
\emph{violating stretch} if $\sigma$ is smaller than the threshold
$t$:
\[
  \sigma \ngeq t
\]

The \emph{violation of a decision variable}, say $\Col{v}$ for vertex
$v$ with $\StretchVertex{v} = \Tuple{\ell,r,c,\sigma}$, where
$\CurrentValue{\Col{v}} = c$, is defined as follows:
\begin{equation*}
  \VarViolation{\Col{v}} =
  \begin{cases}
    0 & \text{if~~}
        v \notin \Set{\ell,r} \\
    0 & \text{if~~}
        v \in \Set{\ell,r} \land \sigma \geq t \land \sigma-\Val{v} \ngeq t \\
    \Val{v} & \text{if~~}
        v \in \Set{\ell,r} \land \sigma \geq t \land \sigma-\Val{v} \geq t \\
    1 & \text{if~~}
        v \in \Set{\ell,r} \land \sigma \ngeq t
  \end{cases}
\end{equation*}
The variable violation is zero if $v$ is currently either not at the
border (leftmost or rightmost element) of its colour stretch (so that
flipping its colour would break its current stretch into \emph{three}
stretches) or at the border of a non-violating colour stretch that
would become violating upon losing $v$.  The variable violation is
positive if $v$ is currently at the border of a colour stretch that is
either non-violating and would remain so upon losing $v$ (so that it
can contribute $\Val{v}$ to the value sum of the adjacent colour
stretch, if any) or violating (so that there is an incentive to drop
$v$ and eventually eliminate this stretch).

The \emph{violation of the constraint} is the current number of
violating colour stretches:
\begin{equation*}
  \Violation =
  \sum_{\Tuple{\_,\_,\_,\sigma} \in \Stretch} \HigherOrder{\sigma \ngeq t}
\end{equation*}
The constraint violation is zero if there currently is no violating
colour stretch.

Let us now measure the additive impact on the constraint violation of
an assignment move $\Col{v} \IsAssigned d$, with $\StretchVertex{v} =
\Tuple{\ell,r,c,\sigma}$.  For simplicity of notation, we assume $d
\neq c$.  Let the colour stretch to the left be
$\Tuple{\ell',r',c',\sigma'}$, with $\Pred{\ell} = r'$ if it exists,
and $\Tuple{\bot,\bot,\gamma,+\infty}$ otherwise, with $\gamma \notin
\Colours$.  Similarly, let the colour stretch to the right be
$\Tuple{\ell'',r'',c'',\sigma''}$, with $\Succ{r} = \ell''$ if it
exists, and $\Tuple{\bot,\bot,\gamma,+\infty}$ otherwise, with $\gamma
\notin \Colours$.  The \emph{assignment delta} function is defined as
follows:
\begin{equation*}
  \begin{array}{l}
    \Probe{\Col{v} \IsAssigned d} \\
    =
    \begin{cases}
      \begin{array}{l}
        -~ \HigherOrder{\sigma' \ngeq t \land
              \HigherOrder{c'=d} \cdot \sigma'
              + \sigma
              + \HigherOrder{c''=d} \cdot \sigma'' \geq t} \\
        -~ \HigherOrder{\sigma \ngeq t} \\
        -~ \HigherOrder{\sigma'' \ngeq t \land
              \HigherOrder{c'=d} \cdot \sigma'
              + \sigma
              + \HigherOrder{c''=d} \cdot \sigma'' \geq t}
      \end{array}
      & \text{if~~} \ell = v = r
      \\
      \HigherOrder{\sigma \geq t \land \sigma-\Val{v} \ngeq t}
      - \HigherOrder{\sigma' \ngeq t \land \sigma'+\Val{v} \geq t}
      & \text{if~~} \ell = v \prec r \land d = c'
      \\
      \HigherOrder{\Val{v} \ngeq t} +
      \HigherOrder{\sigma \geq t \land \sigma-\Val{v} \ngeq t}
      & \text{if~~} \ell = v \prec r \land d \neq c'
      \\
      \text{(analogous to the two previous cases)}
      & \text{if~~} \ell \prec v = r
      \\
      2
      & \text{if~~} \ell \prec v \prec r \land \sigma \ngeq t
      \\
        \HigherOrder{\displaystyle\sum_{\ell \preceq i \prec v} \Val{i} \ngeq t}
      + \HigherOrder{\Val{v} \ngeq t}
      + \HigherOrder{\displaystyle\sum_{v \prec i \preceq r} \Val{i} \ngeq t}
      & \text{if~~} \ell \prec v \prec r \land \sigma \geq t
    \end{cases}
  \end{array}
\end{equation*}
An assignment move on vertex $v$ can be differentially probed in time
at worst linear in the number $\Cardinality{V}$ of vertices.
This is so only in the last case and when there currently is only one
colour stretch.  In all other cases, differential probing takes
constant time.

It is advisable to use a neighbourhood where vertices at the border of
a stretch are re-coloured using a currently unused colour or the
colour of an adjacent connected component.

A swap move $\Col{v} :=: \Col{w}$, where vertices $v$ and $w$ exchange
their colours, is the sequential composition of the two assignment
moves $\Col{v} \IsAssigned \CurrentValue{\Col{w}}$ and $\Col{w}
\IsAssigned \CurrentValue{\Col{v}}$.  The \emph{swap delta} is the sum
of the assignment deltas for these two moves (upon incrementally
making the first move), and there is no asymptotically faster way to
compute this delta, as at most one of these two moves has the
worst-case probing complexity that depends on the number of vertices.

We omit the rather clerical code for achieving incrementality.

\paragraph{Hard Constraint.}
If the $\StretchSum$ constraint is considered implicitly, then it can
be satisfied cheaply in the start assignment, by greedily colouring
the vertices from left to right so that every stretch has an adequate
sum.  However, it may be impossible to maintain this constraint as
satisfied upon every move, even when only considering moves that
re-colour one or more vertices at the border of a stretch to the
colour of an adjacent stretch.

\subsubsection{Propagator}
\label{sect:stretchSum:cp}

A checker for the $\StretchSum(G,\Colour,\Value,\RelOp,t)$ constraint
can be elegantly modelled using the following deterministic finite
automaton (DFA) with a counter and predicates on the transitions:
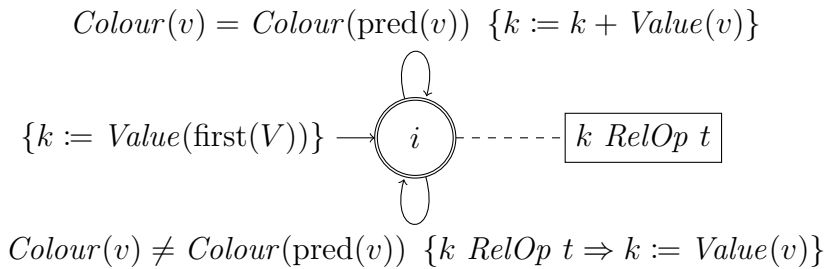
\begin{center}
  \begin{tikzpicture}[node distance=30mm]
    \node[initial,accepting,initial text=\Eq,initial distance=5mm,state] (A) {$i$};
    \node[rectangle,draw] (B) [right of=A] {$k ~\RelOp~ t$};
    \path
    (A)
    edge[loop above] node{$\Arc{\Col{v} = \Col{\Pred{v}}}
      {k \IsAssigned k + \Val{v}}$} (A)
    edge[loop below] node{$\Arc{\Col{v} \neq \Col{\Pred{v}}}
      {k ~\RelOp~ t \Implies k \IsAssigned \Val{v}}$} (A);
    \draw[dashed] (A) to (B);
  \end{tikzpicture}
\end{center}
There is a unique state, called $i$: it is the start state (as marked
by a transition coming in from nowhere) and an accepting state (as
marked by the double circle).  There is a unique counter, called $k$:
it is initialised on the start transition to the value of the leftmost
vertex; without loss of generality, we assume $V \neq \EmptySet$.  The
transitions are made on the truth values of predicates on a current
vertex $v$ and its predecessor.  The counter evolves on each such
transition, unconditionally in the top transition (because $v$ is in
the colour stretch of its predecessor) and conditionally in the bottom
transition (because $v$ starts a new colour stretch).  If a
conditional transition fails, then a transition is made to an implicit
failure state, which has only self-looping transitions.  Upon
processing the rightmost vertex, the accepting state is actually
accepting only if the attached \emph{acceptance condition} (drawn in a
box connected to it by a dashed line) is satisfied: it ensures that
the last stretch has an adequate sum.

Such automata were introduced in~\cite{automaton}.  Assume the
following constraints, called \emph{signature constraints}, for new
Boolean decision variables $\Sig{v}$, called \emph{signature
  variables}, with initial domain $\Set{0,1}$:
\[
  \forall v \in V \setminus \Set{\First{V}} :
  \Col{v} = \Col{\Pred{v}} \Iff \Sig{v} = 1
\]
We can feed these $\Cardinality{V \setminus \Set{\First{V}}}$
signature variables into a DFA with a counter, just like the one
above, except that the top predicate is replaced by alphabet symbol
$1$, and the bottom one by $0$.  It is shown in~\cite{automaton} how
to prototype rapidly a propagator from such a constraint checker using
the $\Automaton$ constraint.  In this case, it probably does not
achieve domain consistency, but one can try and boost its propagation
using our ideas in~\cite{ASTRA:ICTAI13:implied}.

\subsection{Balanced Workload}
\label{sect:balanced}

The $\Balanced(G,\Colour,\Value,\mu,\Delta)$ constraint holds if and
only if the sums of the given integer values under $\Value$ of the
vertices having the same colour under $\Colour$ are balanced, in the
sense of having the (possibly unknown, and not necessarily integer)
value $\mu$ as average and having discrepancies to $\mu$ that do not
exceed the (possibly unknown) integer threshold $\Delta$.  If $\Delta$
is not given, then it may appear in the objective function, towards
being minimised.  Formally, the constraint can be decomposed into the
following conjunction:
\begin{equation}\label{balanced:sem}
  \forall i \in \Colours :
  X[i] = \sum_{v \in V} (\HigherOrder{\Col{v}=i} \cdot \Val{v})
  \land \Gamma(X,\mu,\Delta)
\end{equation}
where constraint $\Gamma$ is either $\Spread$ or $\Deviation$, thereby
giving a concrete definition to the used abstract concept of
\emph{discrepancy}:
\begin{itemize}
\item The $\Spread(X,\Delta,\mu)$ constraint~\cite{Pesant:spread}
  holds if and only if the $n$ integer variables $X[i]$ have the
  (possibly unknown, and not necessarily integer) value $\mu$ as
  average and the sum of the squared differences $(n \cdot X[i] - n
  \cdot \mu)^2$ does not exceed the (possibly unknown) integer
  threshold $\Delta$.
\item The $\Deviation(X,\Delta,\mu)$
  constraint~\cite{Schaus:deviation} holds if and only if the $n$
  integer variables $X[i]$ have the (possibly unknown, and not
  necessarily integer) value $\mu$ as average and the sum of the
  deviations $\AbsValue{n \cdot X[i] - n \cdot \mu}$ does not exceed
  the (possibly unknown) integer threshold $\Delta$.
\end{itemize}
The multiplications by $n$ in the definitions of discrepancy lift all
reasoning to integer domains even when $\mu$ is not an integer, as
$\sum_{i=1}^n X[i] = n \cdot \mu$ and the $X[i]$ are integer
variables: the integer threshold $\Delta$ has to be calibrated
accordingly.  One could also use the
$\Constraint{Range}(X,\RelOp,\Delta)$ constraint~\cite{GC-catalogue},
which holds if and only if $\max(X) + 1 - \min(X) ~\RelOp~ \Delta$,
but we do not pursue this option further.

In airspace sectorisation, the workload of each sector must be within
some given imbalance factor of the average across all sectors.  Hence
one would take $\Value$ as the $\workloadfun$ function of
Section~\ref{sect:math}.  Recall that we only consider additive
workloads, such as monitoring workload and conflict workload, but no
non-additive workloads, such as coordination workload.  Note that
$\mu$ is known when the number $n$ of sectors (and thus colours) is
imposed:
\begin{equation}\label{atm:mu}
  \mu = \frac{\sum_{v \in V} \workloadfun(v)}{n}
\end{equation}

The workload balancing constraint is a soft constraint in our prior
work on airspace sectorisation under stochastic local
search~\cite{ASTRA:ATM13}, but its concept of discrepancy is defined
in terms of a \emph{ratio} rather than a difference with the average
$\mu$, namely $X[i] / \mu \leq 1 + \Delta$ for each $i$ (in the
experiments, $\Delta=0.05$ was used).  This concept of discrepancy is
less related to standard concepts in statistics.

A workload balancing constraint is also used in our prior work on
airspace sectorisation under systematic search~\cite{Jaegare:MSc11},
but again under a different concept of discrepancy with the average
$\mu$, namely $1 - \Delta \leq X[i] / \mu \leq 1 + \Delta$ for each
$i$ (in the experiments, $\Delta=0.25$ was used).  The constraint is
enforced not by a propagator, but by setting the domains of the $X[i]$
integer decision variables to $\Set{\Ceiling{(1-\Delta) \cdot
    \mu},\dots,\Floor{(1+\Delta) \cdot \mu}}$, which is possible when
$\Delta$ is given as $\mu$ is a known constant by~(\ref{atm:mu}).
This modelling trick cannot be applied with the definitions of
discrepancy for $\Spread$ and $\Deviation$.

\subsubsection{Violation and Differentiation Functions}
\label{sect:BW:cbls}

We here handle the $\Balanced$ constraint for the case $\Gamma =
\Deviation$.  Handling the case $\Gamma = \Spread$ can be done using
the same ideas.  We assume $\Colours = \Set{1,2,\dots,n}$.

\paragraph{Soft Constraint.}
If the $\Balanced(G,\Colour,\Value,\mu,\Delta)$ constraint is
considered explicitly, then we proceed as follows.  For simplicity of
notation, we assume $\Delta$ and $\mu$ are given, and that $\mu$ is an
integer.  Relaxing these assumptions can be done using the same ideas
as below.  The multiplications by the number $n$ of colours in the
definition of the underlying $\Deviation$ constraint can then be
eliminated, upon dividing $\Delta$ by $n^2$, giving the following
simplified semantics of $\Balanced$:
\begin{equation}\label{deviation:sem:part1}
  \sum_{i=1}^n X[i] = n \cdot \mu
\end{equation} 
and
\begin{equation}\label{deviation:sem:part2}
  \sum_{i=1}^n \AbsValue{X[i] - \mu} \leq \Delta
\end{equation}
where
\begin{equation}\label{balanced:sem:part1}
  \forall i \in \Set{1,\dots,n} :
  X[i] = \sum_{v \in V} (\HigherOrder{\Col{v}=i} \cdot \Val{v})
\end{equation}
Note that~(\ref{balanced:sem:part1}) implies
\[
  \sum_{i=1}^n X[i] = \sum_{v \in V} \Val{v}
\]
so that, using~(\ref{deviation:sem:part1}), we must have
\begin{equation}\label{balanced:mu}
  \mu = \frac{\sum_{v \in V} \Val{v}}{n}
\end{equation}
similarly to~(\ref{atm:mu}).  From now on, we assume the given
$\Value$ and $\mu$ satisfy~(\ref{balanced:mu}), so that we need not
reason about~(\ref{deviation:sem:part1}), as it is then surely
satisfied, because implied when~(\ref{balanced:sem:part1}) is
satisfied.  Formula~(\ref{balanced:sem:part1}) itself defines the
auxiliary decision variables $X[i]$, so it suffices to set it up as a
set of invariants~\cite{Comet}.  In conclusion, we only need to deal
with formula~(\ref{deviation:sem:part2}).

The \emph{violation of a decision variable}, say $\Col{v}$ for vertex
$v$, is the same as the violation of the auxiliary decision variable
$X[\CurrentValue{\Col{v}}]$, which is functionally dependent on
$\Col{v}$ under~(\ref{balanced:sem:part1}).  This violation is the
current deviation from $\mu$ of the value sum for the current colour
of $v$:
\begin{equation*}
  \VarViolation{\Col{v}}
  = \VarViolation{X[\CurrentValue{\Col{v}}]}
  = \AbsValue{\CurrentValue{X[\CurrentValue{\Col{v}}]} - \mu}
\end{equation*}
The variable violation is zero if $v$ currently has a colour $i$ whose
value sum $X[i]$ is equal to $\mu$, and thus contributes nothing to
the total deviation for all colours.

The \emph{violation of the constraint} is the excess, if any, over
$\Delta$ of the sum of the current deviations from $\mu$ of the value
sums for all colours.  We need not initialise and maintain any
internal data structure for this purpose, as each auxiliary decision
variable $X[i]$ contains the current value sum for colour $i$, and its
variable violation is the current deviation from $\mu$ of the value
sum for colour $i$.  Hence the constraint violation is defined as
follows:
\begin{equation}\label{balanced:cons:vio}
  \Violation
  = \max\left( \sum_{i=1}^n \VarViolation{X[i]} - \Delta,~ 0 \right)
\end{equation}
The constraint violation is zero if the total deviation for all
colours currently does not exceed $\Delta$.

The impact on the constraint violation of an assignment move $\Col{v}
\IsAssigned c$ is measured as follows.  We assume
$\CurrentValue{\Col{v}} = d$.  Let $s$ be the current value of the
first argument of the maximisation expression
in~(\ref{balanced:cons:vio}), and let $s'$ be the same value except
for colours $c$ and $d$:
\begin{equation*}
  s'
  = \sum_{\substack{i=1 \\ i \notin \Set{c,d}}}^n \VarViolation{X[i]}
  - \Delta
  = s
  - \AbsValue{\CurrentValue{X[c]}-\mu}
  - \AbsValue{\CurrentValue{X[d]}-\mu}
\end{equation*}
We get the following \emph{assignment delta} function:
\begin{equation*}
  \begin{array}{l}
    \Probe{\Col{v} \IsAssigned c} \\
    = \max(s'
           + \AbsValue{\CurrentValue{X[c]}+\Val{v}-\mu}
           + \AbsValue{\CurrentValue{X[d]}-\Val{v}-\mu},~ 0)
    - \max(s,~ 0)
  \end{array}
\end{equation*}
An assignment move can be differentially probed in constant time.

A swap move $\Col{v} :=: \Col{w}$, where vertices $v$ and $w$ exchange
their colours, is the sequential composition of the two assignment
moves $\Col{v} \IsAssigned \CurrentValue{\Col{w}}$ and $\Col{w}
\IsAssigned \CurrentValue{\Col{v}}$.  The \emph{swap delta} is the sum
of the assignment deltas for these two moves (upon incrementally
making the first move), and there is no asymptotically faster way to
compute this delta, as probing an assignment move takes constant time.

We omit the rather clerical code for achieving incrementality.

\paragraph{Hard Constraint.}
It is very difficult to consider the $\Balanced$ constraint
implicitly, as it is not obvious how to satisfy it cheaply in the
start assignment and under what probed moves to maintain it satisfied.

\subsubsection{Propagator}
\label{sect:BW:cp}

Due to the decomposition~(\ref{balanced:sem}) of the $\Balanced$
constraint using either the $\Spread$ or the $\Deviation$ constraint,
we refer to~\cite{Pesant:spread,Schaus:deviation} for propagators for
these constraints.  Bounds consistency can be achieved in $\Oh{n \cdot
  \log n}$ time and $\Oh{n}$ time, respectively.  The remaining part
of~(\ref{balanced:sem}), namely
\begin{equation*}
  \forall i \in \Colours :
  X[i] = \sum_{v \in V}
  \left( \HigherOrder{\Col{v}=i} \cdot \Val{v} \right)
\end{equation*}
can be further decomposed as follows, using reification and a 2D
matrix of new Boolean decision variables $B[i,v]$:
\begin{gather*}
  \forall i \in \Colours :
  \forall v \in V : \Reifies{\Col{v}=i}{B[i,v]=1} \\
  \land~
  \forall i \in \Colours :
  \Linear(\Value,B[i,*],=,X[i])
\end{gather*}
The propagation of reified constraints and the $\Linear$ constraint
are standard features of every constraint programming solver.  We
doubt the fixpoint of these propagators achieves domain consistency,
but the development of a custom propagator for $\Balanced$ along the
lines of our~\cite{ASTRA:CP13:convex} is future work.

\subsection{Bounded Workload}
\label{sect:bounded}

The $\Bounded(G,\Colour,\Value,\RelOp,t)$ constraint, with $\RelOp \in
\Set{\leq,<,=,\neq,>,\geq}$, holds if and only if every sum of the
given integer values under $\Value$ of the vertices having the same
colour under $\Colour$ is in relation $\RelOp$ with threshold $t$,
whose value is given.  Formally, the constraint can be decomposed into
the following conjunction:
\begin{equation*}
  \forall i \in \Colours :
  \sum_{v \in V} (\HigherOrder{\Col{v}=i} \cdot \Val{v})
  ~\RelOp~ t
\end{equation*}
This can be further decomposed as follows, using reification and a 2D
matrix of new Boolean decision variables $B[i,v]$:
\begin{gather*}
  \forall i \in \Colours :
  \forall v \in V : \Reifies{\Col{v}=i}{B[i,v]=1} \\
  \land~
  \forall i \in \Colours :
  \Linear(\Value,B[i,*],\RelOp,t)
\end{gather*}
The handling of reified constraints and the $\Linear$ constraint are
standard features of every constraint programming solver, so we need
not develop any new propagators or violation and differentiation
functions.  We doubt the fixpoint of these propagators achieves domain
consistency, but the development of a custom propagator for $\Bounded$
along the lines of our~\cite{ASTRA:CP13:convex} is future work.

In airspace sectorisation, the workload of each sector must not exceed
some upper bound.  Hence one would take $\Value$ as the $\workloadfun$
function of Section~\ref{sect:math}, and $\RelOp$ as $\leq$.  Recall
that we only consider additive workloads, such as monitoring workload
and conflict workload, but no non-additive workloads, such as
coordination workload.

We did not consider a bounded-workload constraint in our prior work on
airspace sectorisation under stochastic local
search~\cite{ASTRA:ATM13} and systematic search~\cite{Jaegare:MSc11},
but such a constraint has been considered by others, as discussed in
our survey~\cite{ASTRA:sectorisation:survey}.

\subsection{Balanced Size}

In airspace sectorisation, the size of each sector must be within some
given imbalance factor of the average across all sectors.  Hence one
can use the $\Balanced$ constraint of Section~\ref{sect:balanced},
with $\Value$ as the $\Volume$ function of Section~\ref{sect:math}.

We did not consider a balanced-size constraint in our prior work on
airspace sectorisation under stochastic local
search~\cite{ASTRA:ATM13} and systematic search~\cite{Jaegare:MSc11},
but such a constraint has been considered by others, as discussed in
our survey~\cite{ASTRA:sectorisation:survey}.

\subsection{Minimum Distance: No Border Vertices in Stretches}
\label{sect:nonBorder}

Let $P$ be the sequence of vertices of a simple path in graph $G =
\Tuple{V,E}$, plus the special vertex $\bot$ at the beginning and at
the end.  The $\NonBorder(G,\Colour,P)$ constraint holds if and only
if all vertices of all stretches of the projection, denoted by
$\Colour(P)$, of $\Colour$ onto the vertices of $P$ (and in the vertex
ordering of $P$) only have adjacent vertices outside $P$ of the same
colour:
\begin{equation*}
  \begin{array}{c}
    \forall \ell \preceq r \in P \setminus \Sequence{\bot} :
    \Stretch(\Colour(P),\ell,r) \\
    \Implies
    \forall \ell \preceq v \preceq r \in P :
    \forall w \in \Adjacent{v} \setminus P : \Col{w} = \Col{v}
  \end{array}
\end{equation*}
This constraint is trivially satisfied when the graph $G$ is induced
by a one-dimensional geometry, as every vertex in $P$ then has
\emph{no} adjacent vertices outside $P$.  Note that there is no limit
on the number of stretches per colour.

In airspace sectorisation, each existing trajectory must be inside
each sector by a minimum distance (say ten nautical miles), so that
conflict management is entirely local to sectors.  If the airspace is
originally divided into same-sized regions whose diameter is (at
least) that minimum distance, then the border regions of each sector
can only serve as sector entry and exit regions for all flights.
Hence one could use a $\NonBorder$ constraint for each flight $f$, by
setting $P$ to its sequence of visited regions:
\[
  P = 
  \Sequence{\bot}
  \Union
  \Sequence{v_i \mid \Tuple{v_i,\_,\_} \in \Plan(f)}
  \Union
  \Sequence{\bot}
\]
In our prior work on airspace sectorisation under systematic
search~\cite{Jaegare:MSc11}, we did not need such a minimum-distance
constraint, because any regions that cannot serve as border regions of
sectors were pre-aggregated into AFBs.  In our prior work on airspace
sectorisation under stochastic local search~\cite{ASTRA:ATM13}, we
initially used the same pre-aggregation into AFBs in order to avoid
having such a minimum-distance constraint, but it then turned out that
the $\Compact$ constraint (see Section~\ref{sect:compact}) is very
hard to satisfy when AFBs are used, so we had concluded that a
minimum-distance constraint would be necessary after all, upon
switching off the pre-aggregation into AFBs, but we did not design
such a constraint.

\subsubsection{Violation and Differentiation Functions}

\paragraph{Soft Constraint.}
If the $\NonBorder$ constraint is considered explicitly, then we
proceed as follows, without needing any internal datastructures.

The \emph{violation of a decision variable}, say $\Col{v}$ for vertex
$v$, is defined as follows:
\begin{equation*}
  \VarViolation{\Col{v}} =
  \begin{cases}
    0 & \text{if~~} v \notin P \\
    \displaystyle\sum_{w \in \Adjacent{v} \setminus P}
    \HigherOrder{\CurrentValue{\Col{w}}
      \neq \CurrentValue{\Col{v}}}
    & \text{if~~} v \in P
  \end{cases}
\end{equation*}
The variable violation is zero if $v$ is not in $P$ or has no adjacent
vertices outside $P$ that currently have a different colour.

The \emph{violation of the constraint} is the current sum of
the violations of its variables:
\begin{equation*}
  \Violation = \sum_{v \in P} \VarViolation{\Col{v}}
\end{equation*}
The constraint violation is zero if the constraint is satisfied.

The additive impact on the constraint violation of an assignment move
$\Col{v} \IsAssigned c$ for vertex $v \in P$ is measured by the
following \emph{assignment delta} function:
\begin{equation*}
  \Probe{\Col{v} \IsAssigned c} =
  \sum_{w \in \Adjacent{v} \setminus P}
  \left(
    \HigherOrder{\CurrentValue{\Col{w}} \neq c}
    -
    \HigherOrder{\CurrentValue{\Col{w}} \neq \CurrentValue{\Col{v}}}
  \right)
\end{equation*}
An assignment move on vertex $v$ can be differentially probed in time
linear in the degree of $v$ in $G$.

It is advisable to use a neighbourhood where vertices at the border of
a stretch of $\Colour(P)$ are re-coloured using a currently unused
colour or the colour of an adjacent vertex outside $P$.

A swap move $\Col{v} :=: \Col{w}$, where vertices $v$ and $w$ exchange
their colours, is the sequential composition of the two assignment
moves $\Col{v} \IsAssigned \CurrentValue{\Col{w}}$ and $\Col{w}
\IsAssigned \CurrentValue{\Col{v}}$.  The \emph{swap delta} is the sum
of the assignment deltas for these two moves (upon incrementally
making the first move), and there is no asymptotically faster way to
compute this delta, as the complexity of probing an assignment move
does not depend on the number of vertices.

We omit the rather clerical code for achieving incrementality.

\paragraph{Hard Constraint.}
It is very difficult to consider the $\NonBorder$ constraint
implicitly, as it is not obvious how to satisfy it cheaply in the
start assignment and under what probed moves to maintain it satisfied.

\subsubsection{Propagator}

Designing a propagator for the $\NonBorder$ constraint is left as
future work.  Note that a propagator cannot even be prototyped using
the $\Automaton$ constraint, as the constraint is not only on the
$\Colour(P)$ sequence of decision variables, but also on several
$\Colour(\Adjacent{i})$ sequences of decision variables, the set of
such vertices $i$ not being known up front.  A different kind of
automaton (possibly with counters) would have to be imagined, and
supported with a new constraint like $\Automaton$, in order to get a
propagator for $\NonBorder$.

\section{Conclusion}
\label{sect:concl}

Airspace sectorisation provides a partition of a given airspace into
sectors, subject to geometric constraints and workload constraints, so
that some cost metric is minimised.  We have studied the constraints
that arise in airspace sectorisation, giving for each constraint an
analysis of what algorithms and properties are required under
systematic search and stochastic local search.

\subsection{Discussion}

The formal semantics or decompositions of several constraints exhibit
a pattern that we had not encountered before: the decision variables
of the problem functionally determine auxiliary decision variables,
which are the arguments of well-known constraints.  For example, the
auxiliary variables $X[i]$ of the decomposition~(\ref{balanced:sem})
of the $\Balanced$ constraint in Section~\ref{sect:balanced} are
functionally determined by the $\Col{v}$ variables, but it is the
$X[i]$ that are constrained by the well-known $\Spread$ or
$\Deviation$ constraints.  This level of indirection makes propagation
difficult, but apparently often occurs in real-life problems:
overcoming this is a major research challenge for all optimisation
technologies, and thus beyond the scope of this report.

We were surprised how little propagation is possible for some airspace
sectorisation constraints, especially the $\Compact$ and $\Connected$
constraints: this explains why our prior work using systematic
search~\cite{Jaegare:MSc11} was not as successful as we had hoped.  In
retrospect, we advocate achieving sector compactness by using also a
constraint on \emph{maximum} dwell time, that is
$\StretchSum(\Colour,\Value,\leq,t)$ for some suitable upper bound $t$
on dwell time.  We believe this would trigger more propagation before
most of the decision variables have singleton domains (see
Section~\ref{sect:compact:cp}).  However, we do not know whether
airspace sectorisation experts would be willing to quantify the upper
bound $t$.

We have noticed that the violation and delta functions for stochastic
local search are sometimes hard to design in a problem-independent
fashion, towards making the constraints highly reusable.  Indeed, one
can often imagine several pairs of measures for variable and
constraint violation (if not constraint semantics), each pair needing
a rather different mathematical and algorithmic apparatus for
incremental maintenance and differential probing, but the appropriate
pair may depend on the actual problem or on the neighbourhood chosen
for search.  It thus seems that some constraints need to be offered in
several incarnations, as witnessed with the $\Compact$ constraint and
with other choices discussed in the literature we have pointed to:
there is no reason to believe that one set of choices dominates all
others.  The same phenomenon occurs with propagators for systematic
search, where different levels of consistency can be aimed at, and
where different time and space complexity trade-offs exist even for
propagators achieving the same level of consistency.

\subsection{Related Work}

For each constraint in Section~\ref{sect:geo}, we have given a
discussion of all similar constraints that we are aware of, including
literature pointers.

\subsection{Future Work}

Coordination workload is not additive when aggregating the workload of
a sector from the workloads of its constituent regions.  We have only
handled the monitoring and conflict workloads in this report, because
they are additive and thus easier to reason with, and we have assumed
they are combined into a single value.  This work needs to be extended
to coordination workloads, and possibly to a separate handling of all
three kinds of workload.

For time reasons, we have not tackled all the constraints arising in
airspace sectorisation (identified in Section~\ref{sect:secto}), so
the few remaining constraints need to be discussed in equal depth.

The constraints should also be tackled under the set covering
approach, as we have assumed here the graph colouring approach for all
the covered constraints.  The representations of the two approaches
are dual (in the sense that the set covering approach represents the
inverse of the total function, of the graph colouring approach, from
regions to sectors).  It may turn out to be beneficial to switch the
representation for \emph{some} constraints, if not \emph{always} to
use both representations at the same time, and to channel between
them.

The completion of our line of work, based on constraint programming as
a combinatorial problem solving technology, will show that \emph{all}
the constraints can be used in the process of \emph{computing} a
sectorisation, rather than only using some and then evaluating the
\emph{results} of a sectorisation algorithm according to the other
constraints.

\bibliographystyle{abbrv}
\bibliography{sectorisation,atm-cp}

\begin{thebibliography}{10}

\bibitem{cumulative}
A.~Aggoun and N.~Beldiceanu.
\newblock Extending {CHIP} in order to solve complex scheduling and placement
  problems.
\newblock {\em Mathematical and Computer Modelling}, 17(7):57--73, 1993.

\bibitem{ASTRA:KER:survey}
C.~Allignol, N.~Barnier, P.~Flener, and J.~Pearson.
\newblock Constraint programming for air traffic management: {A} survey.
\newblock {\em The Knowledge Engineering Review}, 27(3):361--392, September
  2012.

\bibitem{automaton}
N.~Beldiceanu, M.~Carlsson, and T.~Petit.
\newblock Deriving filtering algorithms from constraint checkers.
\newblock In M.~Wallace, editor, {\em CP 2004, the 10th International
  Conference on Principles and Practice of Constraint Programming}, volume 3258
  of {\em LNCS}, pages 107--122. Springer-Verlag, 2004.

\bibitem{GC-catalogue}
N.~Beldiceanu, M.~Carlsson, and J.-X. Rampon.
\newblock Global constraint catalogue: {P}ast, present, and future.
\newblock {\em Constraints}, 12(1):21--62, March 2007.
\newblock The current working version of the catalogue is at
  \url{http://www.emn.fr/z-info/sdemasse/aux/doc/catalog.pdf}.

\bibitem{Beldiceanu:graphProps:bounds}
N.~Beldiceanu, T.~Petit, and G.~Rochart.
\newblock Bounds of graph parameters for global constraints.
\newblock {\em RAIRO -- Operations Research}, 40(4):327--353, October--December
  2006.

\bibitem{ASTRA:sectorisation:survey}
P.~Flener and J.~Pearson.
\newblock Automatic airspace sectorisation: {A} survey.
\newblock Technical Report 1311.0653, Computing Research Repository, November
  2013.
\newblock Available at \url{http://arxiv.org/abs/1311.0653}.

\bibitem{ASTRA:ICTAI13:implied}
M.~A. Francisco~Rodr\'iguez, P.~Flener, and J.~Pearson.
\newblock Generation of implied constraints for automaton-induced
  decompositions.
\newblock In E.~Gr\'egoire and B.~Mazure, editors, {\em ICTAI/CSP 2013, the
  special track on SAT and CSP technologies of the 25th IEEE International
  Conference on Tools with Artificial Intelligence}. IEEE Computer Society,
  2013.

\bibitem{ASTRA:JoH}
J.~He, P.~Flener, and J.~Pearson.
\newblock Underestimating the cost of a soft constraint is dangerous:
  {R}evisiting the edit-distance based {SoftRegular} constraint.
\newblock {\em Journal of Heuristics}, 19(5):729--756, October 2013.

\bibitem{Hooker:integrated}
J.~N. Hooker.
\newblock {\em Integrated Methods for Optimization}.
\newblock Springer-Verlag, second edition, 2011.

\bibitem{Hoos:SLS}
H.~H. Hoos and T.~St{\"u}tzle.
\newblock {\em Stochastic Local Search: {F}oundations {\&} Applications}.
\newblock Elsevier / Morgan Kaufmann, 2004.

\bibitem{Jaegare:MSc11}
P.~J\"agare.
\newblock Airspace sectorisation using constraint programming.
\newblock Master's thesis, Uppsala University, Sweden, Report IT 11 021,
  Faculty of Science and Technology, 2011.
\newblock Available at
  \url{http://urn.kb.se/resolve?urn=urn:nbn:se:uu:diva-155783}.

\bibitem{ASTRA:ATM13}
P.~J\"agare, P.~Flener, and J.~Pearson.
\newblock Airspace sectorisation using constraint-based local search.
\newblock In S.~Saunders-Hodge and C.~Meckiff, editors, {\em ATM 2013, the 10th
  USA / Europe Seminar on Air Traffic Management R\&D, Chicago, Illinois, USA},
  2013.
\newblock Best paper of theme: Network and Strategic Traffic Flow Optimization.

\bibitem{Maher:contiguity}
M.~J. Maher.
\newblock Analysis of a global contiguity constraint.
\newblock In {\em Workshop on Rule-Based Constraint Reasoning and Programming,
  held at CP 2002, the 8th International Conference on Principles and Practice
  of Constraint Programming}, 2002.

\bibitem{May:algebraic_topology}
J.~May.
\newblock {\em A Concise Course in Algebraic Topology}.
\newblock Chicago Lectures in Mathematics. University of Chicago Press, 1999.

\bibitem{ASTRA:CP13:convex}
J.-N. Monette, N.~Beldiceanu, P.~Flener, and J.~Pearson.
\newblock A parametric propagator for discretely convex pairs of sum
  constraints.
\newblock In C.~Schulte, editor, {\em CP 2013, the 19th International
  Conference on Principles and Practice of Constraint Programming}, volume 8124
  of {\em LNCS}, pages 529--544. Springer, 2013.

\bibitem{Pesant:spread}
G.~Pesant and J.-C. R{\'e}gin.
\newblock {SPREAD}: {A} balancing constraint based on statistics.
\newblock In P.~van Beek, editor, {\em CP 2005, the 11th International
  Conference on Principles and Practice of Constraint Programming}, volume 3709
  of {\em LNCS}, pages 460--474. Springer, 2005.

\bibitem{alldifferent}
J.-C. R{\'e}gin.
\newblock A filtering algorithm for constraints of difference in {CSP}s.
\newblock In B.~Hayes-Roth and R.~E. Korf, editors, {\em AAAI 1994, the 12th
  (US) National Conference on Artificial Intelligence}, pages 362--367. AAAI
  Press, 1994.

\bibitem{gcc}
J.-C. R{\'e}gin.
\newblock Generalized arc-consistency for global cardinality constraint.
\newblock In D.~Weld and B.~Clancey, editors, {\em AAAI 1996, the 13th (US)
  National Conference on Artificial Intelligence}, pages 209--215. AAAI Press,
  1996.

\bibitem{Schaus:deviation}
P.~Schaus, Y.~Deville, and P.~Dupont.
\newblock Bound-consistent deviation constraint.
\newblock In C.~Bessi\`ere, editor, {\em CP 2007, the 13th International
  Conference on Principles and Practice of Constraint Programming}, volume 4741
  of {\em LNCS}, pages 620--634. Springer, 2007.

\bibitem{element}
P.~Van~Hentenryck and J.-P. Carillon.
\newblock Generality versus specificity: {A}n experience with {AI} and {OR}
  techniques.
\newblock In T.~Mitchell and R.~Smith, editors, {\em AAAI 1988, the 7th (US)
  National Conference on Artificial Intelligence}, pages 660--664. AAAI Press,
  1988.

\bibitem{Comet}
P.~Van~Hentenryck and L.~Michel.
\newblock {\em Constraint-Based Local Search}.
\newblock The MIT Press, 2005.

\bibitem{sphericity}
H.~Wadell.
\newblock Volume, shape, and roundness of quartz particles.
\newblock {\em The Journal of Geology}, 43(3):250--280, April/May 1935.

\end{thebibliography}

\end{document}